%% file: main.tex
\documentclass[]{ailab}
\microtypesetup{expansion=false}

\usepackage{silence}
\usepackage{wrapfig}
\usepackage{tabularx}
\usepackage{textcomp}
\usepackage{stfloats}
\usepackage{url}
\usepackage{verbatim}
\usepackage{titlesec}
\usepackage{tocloft}
\usepackage{adjustbox}
\usepackage{multirow}
\usepackage{pifont}
\usepackage{tikz}
\usepackage{comment}
\usepackage{amsmath,amssymb} %
\usepackage{colortbl}  %
\usepackage{color}
\usepackage{booktabs} 
\usepackage{natbib} 
\setcitestyle{square,comma,numbers,sort&compress}
\usepackage{graphicx}    %
\usepackage{subcaption} 
\usepackage{multirow} %
\usepackage{tcolorbox}
\usepackage{booktabs} %
\usepackage{subcaption} %

\usepackage[dvipsnames]{xcolor}
\usepackage{graphicx}
\usepackage{amsmath, amssymb}
\RequirePackage{xspace}
\makeatletter
\DeclareRobustCommand\onedot{\futurelet\@let@token\@onedot}
\def\@onedot{\ifx\@let@token.\else.\null\fi\xspace}

\usepackage{lineno}

\definecolor{darkblue}{rgb}{0, 0, 0.5}
\hypersetup{colorlinks=true, citecolor=darkblue, linkcolor=darkblue, urlcolor=darkblue}

\usepackage{wrapfig}  %
\usepackage{lipsum}

\newtcbtheorem[number within=section]{exmp}{Prompts}%
{breakable,colback=white!5!white,colframe=black!95!,fonttitle=\bfseries, left=.02in, right=.02in,bottom=.02in, top=.02in}{exmp}

\newtcbtheorem[number within=section]{case_data}{Examples}%
{breakable,
 colback=white!5!white,
 colframe={rgb,255:red,78; green,136; blue,114}, %
 fonttitle=\bfseries,
 left=.02in,
 right=.02in,
 bottom=.02in,
 top=.02in}{case_data}

\definecolor{mygreen}{HTML}{85A490}
\definecolor{myred}{HTML}{C75841}

\definecolor{cotred}{RGB}{200, 30, 30}
\definecolor{ForestGreen}{RGB}{34, 139, 34}

\usepackage[normalem]{ulem}

\makeatother

\definecolor{adptorange}{RGB}{248, 205, 172}
\definecolor{cmpblue}{RGB}{189, 215, 238}
\definecolor{cmpblue}{RGB}{189, 215, 238}

\definecolor{our_red}{RGB}{232,157,160}
\definecolor{our_blue}{RGB}{136,206,230}
\definecolor{our_orange}{RGB}{246,200,168}
\definecolor{our_green}{RGB}{178,211,164}

\definecolor{attn_code0}{RGB}{247,215,200}
\definecolor{attn_code1}{RGB}{238,169,139}
\definecolor{mlp_code0}{RGB}{204,201,221}
\definecolor{mlp_code1}{RGB}{102,95,153}

\definecolor{token_blue}{RGB}{84, 120, 140}

\input{math_commands}

\input{requirements/table}

\usepackage{pifont}

\usepackage{pifont}       %
\usepackage{bbding}       %
\usepackage{fontawesome}
\usepackage{xspace}
\usepackage{float}
\usepackage{enumitem}

\def\vx{{\bm{x}}}

\newlength\savewidth

\newcolumntype{x}[1]{>{\centering\arraybackslash}p{#1pt}}
\newcolumntype{y}[1]{>{\raggedright\arraybackslash}p{#1pt}}
\newcolumntype{z}[1]{>{\raggedleft\arraybackslash}p{#1pt}}

\renewcommand{\paragraph}[1]{\vspace{1mm}\noindent\textbf{#1}}
\usepackage{colortbl}
\usepackage{xcolor}
\usepackage{wrapfig}

\renewcommand{\paragraph}[1]{\vspace{1.25mm}\noindent\textbf{#1}}

\usepackage{algorithm}
\usepackage{listings}

\definecolor{codeblue}{rgb}{0.25, 0.5, 0.5}
\definecolor{codekw}{rgb}{0.35, 0.35, 0.75}
\lstdefinestyle{Pytorch}{
    language = Python,
    backgroundcolor = \color{white},
    basicstyle = \fontsize{9pt}{8pt}\selectfont\ttfamily\bfseries,
    columns = fullflexible,
    aboveskip=1pt,
    belowskip=1pt,
    breaklines = true,
    captionpos = b,
    commentstyle = \color{codeblue},
    keywordstyle = \color{codekw},
}

\definecolor{colSubject}{HTML}{D32F2F}   %
\definecolor{colAction}{HTML}{F57C00}    %
\definecolor{colDetail}{HTML}{388E3C}    %
\definecolor{colSpatial}{HTML}{1976D2}   %
\definecolor{colMood}{HTML}{7B1FA2}      %
\definecolor{colKnow}{HTML}{AFB42B}      %

\usepackage{xcolor}

\usepackage{tikz}
\usetikzlibrary{tikzmark, calc, shadows.blur, shapes.geometric, fit, positioning}
\usepackage{xparse}
\pgfdeclarelayer{background}
\pgfdeclarelayer{floatbox}
\pgfdeclarelayer{foreground}
\pgfsetlayers{background,floatbox,main,foreground}

\newcounter{hcellcount}

\NewDocumentCommand{\hctext}{m}{\csname hctext@#1\endcsname}
\NewDocumentCommand{\sethctext}{mm}{\expandafter\gdef\csname hctext@#1\endcsname{#2}}

\definecolor{scoreRed}{RGB}{200, 0, 0}
\definecolor{grayText}{RGB}{120, 120, 120}

\definecolor{green}{HTML}{009000}
\definecolor{red}{HTML}{ea4335}

\definecolor{cvblue}{rgb}{0.15, 0.45, 0.68}

\usepackage{mathtools}
\usepackage{nicefrac}
\usepackage{microtype}
\usepackage{algpseudocode}
\usepackage{longtable}
\usepackage{ragged2e}

\definecolor{codebg}{HTML}{F7F8FA}
\definecolor{codeframe}{HTML}{D0D7DE}

\usepackage[most]{tcolorbox}
\tcbuselibrary{listings,skins,breakable}

\usepackage[scaled=0.85]{beramono}

\definecolor{codebg}{HTML}{FAFBFC}      %
\definecolor{codeframe}{HTML}{E1E4E8}   %
\definecolor{codekw}{HTML}{0033B3}      %
\definecolor{codecmt}{HTML}{8A8F98}     %
\definecolor{codestr}{HTML}{067D17}     %
\definecolor{codenum}{HTML}{B0B6BC}     %

\newtcblisting{pycodebox}{
  listing engine=listings,
  listing options={
    language=Python,
    basicstyle=\scriptsize\ttfamily,
    keywordstyle=\color{codekw},
    commentstyle=\color{codecmt}\itshape,
    stringstyle=\color{codestr},
    numbers=left,
    numberstyle=\tiny\color{codenum},
    numbersep=8pt,
    xleftmargin=16pt,
    breaklines=true,
    breakatwhitespace=false,
    showstringspaces=false,
    tabsize=4,
    columns=fullflexible,
    keepspaces=true,
    upquote=true,
    aboveskip=2pt,
    belowskip=2pt,
  },
  listing only,
  enhanced,
  colback=codebg,
  colframe=codeframe,
  boxrule=0.4pt,
  arc=3pt,
  left=1mm,
  right=2mm,
  top=1.5mm,
  bottom=1.5mm,
  boxsep=0pt
}

\newtheorem{definition}{Definition}[section]

\newcommand{\dyrope}{RiRoPE\xspace}
\newcommand{\acronym}{RiRoPE\xspace}

\colorlet{dyrfill}{metabg!10}               %
\colorlet{dyrframe}{metabg}                 %
\colorlet{dyrtext}{metabg!80!black}         %
\DeclareRobustCommand{\dyropestep}[1]{%
  \ifmmode
    \begingroup\setlength{\fboxsep}{1.5pt}\setlength{\fboxrule}{0.4pt}%
    \mathord{\fcolorbox{dyrframe}{dyrfill}{$#1$}}\endgroup
  \else
    \begingroup\setlength{\fboxsep}{2pt}\setlength{\fboxrule}{0.4pt}%
    \fcolorbox{dyrframe}{dyrfill}{#1}\endgroup
  \fi}
\DeclareRobustCommand{\dyropecomment}[1]{\textcolor{dyrtext}{#1}}
\DeclareRobustCommand{\dyropeswatch}{%
  \begingroup\setlength{\fboxsep}{1pt}\setlength{\fboxrule}{0.4pt}%
  \fcolorbox{dyrframe}{dyrfill}{\rule{0pt}{1.5ex}\hspace{1.1em}}\endgroup}

\title{Native Position Reindexing for Sparse Attention}

\author[1,*]{Lei Zhu}
\author[1,2,*]{Zexuan Qiu}
\author[1,3,*]{Zhiwei Li}
\author[1]{Zhongwei Wan}
\author[4]{Tong Che}
\author[1]{Huayang Li}
\author[1]{Tian Liang}
\author[1,5]{Hao Gu}
\author[3]{Zhijiang Guo}
\author[2]{Irwin King}
\author[1]{Leo Liang}

\affiliation[1]{Tencent HY Team}
\affiliation[2]{The Chinese University of Hong Kong}
\affiliation[3]{Hong Kong University of Science and Technology (Guangzhou)}
\affiliation[4]{Nvidia Inc.}
\affiliation[5]{Hong Kong University of Science and Technology}

\abstract{

Sparse attention typically inherits positional encodings designed for dense attention, despite changing the context available to each query.
Sparse attention bounds the number of attended keys, but these encodings can retain large original-sequence distances that are poorly covered or unseen during training.
We argue that positional distances should instead be co-designed with sparse attention to improve long-context modeling by allowing long-range interactions to share positional representations with short-range ones.
To realize this co-design, we make position reindexing \emph{native} to sparse-attention training: each query's selected tokens receive consecutive positions in their original order during both training and inference, ignoring the gaps left by unselected tokens.
Reindexing limits relative positional distances to the attention budget rather than the full sequence length.
Unlike inference-time reindexing in StreamingLLM, native training lets the model adapt to these reassigned, query-dependent positions rather than merely tolerate them.
Applying this indexing rule to the commonly used rotary position encoding (RoPE) yields ReIndexing RoPE (\acronym).
However, a technical challenge is that query-dependent rotations hinder reuse of rotated keys, erasing the efficiency gains of sparse attention in naive training implementations.
We address this challenge with structure-aware rotation reuse and fused kernels that preserve tensor-core matrix multiplication without materializing per-pair rotated keys.
With matched OLMo-3-7B continued pretraining, \acronym outperforms a sparse-attention baseline using RoPE with YaRN at 64K by 6.14 RULER points ($74.10 \to 80.24$) and 4.40 HELMET points ($42.97 \to 47.37$).
At 512K---eight times the training length---it retains approximately 90\% of its 64K RULER score without further training or positional retuning.
Native training also outperforms inference-only switching by 21.24 RULER points at 256K.
Our work motivates a shift from inheriting positional encodings to co-designing them with sparse attention, reframing sparsity as not merely a computational constraint but an opportunity for better long-context modeling.
}

\date{\today}
\code{\href{https://github.com/Tencent-Hunyuan/RiRoPE}{https://github.com/Tencent-Hunyuan/RiRoPE}}

\begin{document}
\thispagestyle{firstheader}
\maketitle

\input{sec/1_intro}

\input{sec/2_relatedwork}

\input{sec/3_method}
\input{sec/4_experiments}

\input{sec/5_limitation}

\input{sec/6_conclusion}
\input{sec/acknowledgments}

\bibliographystyle{plainnat}
\bibliography{references}

\clearpage
\appendix
\input{sec/7_appendix}

\end{document}

%% file: math_commands.tex
\usepackage{amsmath,amsfonts,bm}

\def\figref#1{figure~\ref{#1}}

\def\eqref#1{equation~\ref{#1}}

\def\1{\bm{1}}

\def\vzero{{\bm{0}}}
\def\vone{{\bm{1}}}

\def\vk{{\bm{k}}}

\def\vm{{\bm{m}}}

\def\vo{{\bm{o}}}

\def\vq{{\bm{q}}}

\def\vv{{\bm{v}}}

\def\vx{{\bm{x}}}

\def\mA{{\bm{A}}}

\def\mK{{\bm{K}}}

\def\mP{{\bm{P}}}
\def\mQ{{\bm{Q}}}

\def\mV{{\bm{V}}}
\def\mW{{\bm{W}}}

\DeclareMathAlphabet{\mathsfit}{\encodingdefault}{\sfdefault}{m}{sl}
\SetMathAlphabet{\mathsfit}{bold}{\encodingdefault}{\sfdefault}{bx}{n}

\def\gA{{\mathcal{A}}}
\def\gB{{\mathcal{B}}}

\def\gI{{\mathcal{I}}}

\newcommand{\R}{\mathbb{R}}

\usepackage{amsmath,amsfonts,bm}

\def\figref#1{figure~\ref{#1}}

\def\eqref#1{equation~\ref{#1}}

\def\1{\bm{1}}

\def\vzero{{\bm{0}}}
\def\vone{{\bm{1}}}

\def\vk{{\bm{k}}}

\def\vm{{\bm{m}}}

\def\vo{{\bm{o}}}

\def\vq{{\bm{q}}}

\def\vv{{\bm{v}}}

\def\vx{{\bm{x}}}

\def\mA{{\bm{A}}}

\def\mK{{\bm{K}}}

\def\mP{{\bm{P}}}
\def\mQ{{\bm{Q}}}

\def\mV{{\bm{V}}}
\def\mW{{\bm{W}}}

\DeclareMathAlphabet{\mathsfit}{\encodingdefault}{\sfdefault}{m}{sl}
\SetMathAlphabet{\mathsfit}{bold}{\encodingdefault}{\sfdefault}{bx}{n}

\def\gA{{\mathcal{A}}}
\def\gB{{\mathcal{B}}}

\def\gI{{\mathcal{I}}}

%% file: requirements/table.tex
\usepackage{multirow}
\usepackage{diagbox}
\usepackage{makecell}
\usepackage{tabularx}
\usepackage{graphicx}

\usepackage{array}
\usepackage{rotating}

\definecolor{aliceblue}{rgb}{0.94, 0.97, 1.0}
\definecolor{citecolor}{HTML}{0071BC}
\definecolor{linkcolor}{HTML}{ED1C24}
\definecolor{darkgreen}{HTML}{539165}

\makeatletter
\newcommand{\thickhline}{%
 \noalign {\ifnum 0=`}\fi \hrule height 1pt
 \futurelet \reserved@a \@xhline
}
\makeatother

\newcommand{\tablesize}{
  \fontsize{8.6pt}{10pt}\selectfont
}

\newcommand{\tablelargesize}{
  \fontsize{9.5pt}{10pt}\selectfont
}

\definecolor{oursbg}{HTML}{F8EBE8}
\definecolor{groupbg}{HTML}{F1F3F4}

\newcommand{\tbest}[1]{{\rmfamily\bfseries #1}}  %
\newcommand{\tsecond}[1]{\underline{#1}}         %
\newcommand{\tours}[1]{{\rmfamily\bfseries #1}}  %
\newcommand{\tna}{\textcolor{gray!55}{--}}       %

\newcommand{\tref}[1]{\textcolor{black!45}{#1}}

%% file: sec/1_intro.tex
\section{Introduction}
\label{sec:intro}

\begin{figure}[t]
  \centering
  \includegraphics[width=\linewidth]{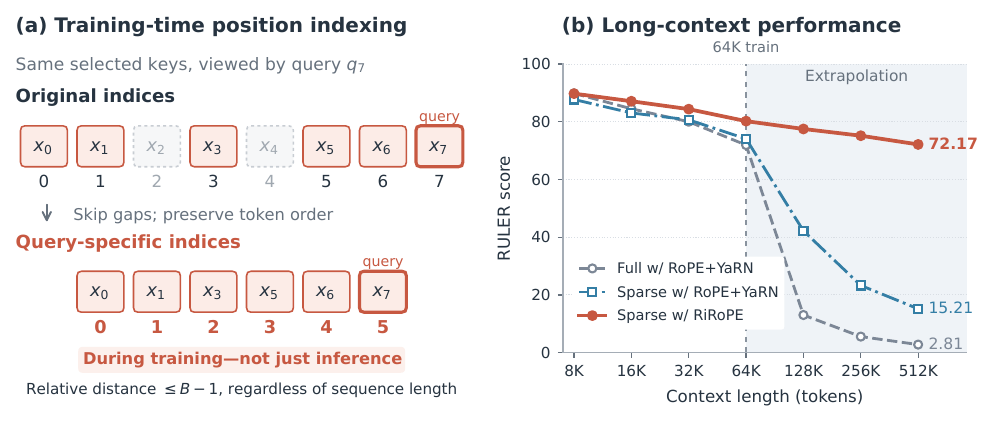}
  \caption{\textbf{Position reindexing for sparse attention and its long-context performance.}
  \textbf{(a)} For one query, the selected keys (colored tokens, including the key corresponding to the query token) receive consecutive indices in their original order, skipping unselected positions (dashed tokens). The model uses these query-specific indices \textbf{during training}, not only at inference; with at most $B$ selected tokens, relative distances stay within $[0,B-1]$ independently of sequence length.
  \textbf{(b)} RULER~\citep{hsieh2024ruler} scores for models with the same continued-pretraining recipe. The dashed line marks the 64K training length; shading denotes evaluation beyond it, without further training or changes to positional-encoding settings.
  }
  \label{fig:teaser}
\end{figure}

Long-context attention faces two distinct length-dependent pressures. As more keys enter the softmax, its entropy grows with the size of the normalization axis, raising the risk of diffuse attention \citep{nakanishi2025ssmax,tan2025stickbreakingattn}.
At the same time, positional encodings must handle relative distances that are rare or unseen during training~\cite{peng2024yarn,chen2023posinterpolation}.
Sparse attention bounds the number of keys normalized over for each query, limiting the first pressure -- entropy inflation as input length increases -- while reducing computation \citep{yuan2025nsa,lu2026moba,lai2026msa}.
Yet it typically leaves the second intact: selected keys retain their original indices, so under Rotary Position Encoding (RoPE) \citep{su2024roformer}, a faraway selected key can still be separated from its query by a large positional distance even when few keys are attended.
The model must therefore still handle relative distances that are poorly covered or unseen during training, which can hinder its use of distant context.
Bounding the attention budget does not by itself bound the positional range; 
we instead seek positional coordinates defined by each query's selected context.

A precedent for such context-dependent positional coordinates comes from streaming inference, where cache eviction leaves only a subset of earlier tokens available for
continued text generation beyond the context window: retained tokens can remain far apart in their original positions.
Streaming methods address the position out-of-range problem by re-indexing the surviving keys consecutively during decoding \citep{xiao2024streamingllm,lee2025infinitehip}.
But this is a post-hoc adaptation of models trained without the re-indexed positions; to learn to use these positions, the model needs to encounter them during training.

We make position reindexing native to sparse-attention training: as illustrated in Figure~\ref{fig:teaser}(a), each query's selected keys receive consecutive indices in their original order during \emph{both training and inference}.
Skipped tokens no longer contribute to the distance between an attended pair; if each query attends to at most $B$ tokens, that distance lies between $0$ and $B-1$, independently of the full sequence length.
The rule strictly generalizes ordinary position indexing: when no tokens are skipped, as for short sequences attended in full within the attention budget, the rule reduces to ordinary position indexing.
Training with these positions lets the model learn to use distances defined by the sparse pattern, rather than tolerating them only at inference.

This reindexing rule changes how selected tokens are positioned, but does not prescribe a particular positional encoding.
We instantiate this rule with RoPE to obtain \textbf{R}e\textbf{I}ndexing \textbf{RoPE} (\acronym), motivated by its widespread use and compatibility with OLMo-3's pretrained weights~\citep{olmo3}.
Continued pretraining of OLMo-3 keeps compute costs manageable, while its fully open training pipeline supports transparent, controlled experiments.
Efficient training is challenging: unlike decoding, which handles one query at a time, sparse-attention training processes many queries in parallel, each with a potentially different selection. The same key may therefore require different positional indices for different queries in one forward pass.
With RoPE, a naive implementation rotates each key separately for every query that selects it. Materializing these rotations incurs substantial memory overhead; fusing them avoids this overhead but does not by itself restore efficient matrix multiplication.
We instead exploit structure already present in sparse attention. For block-sparse multi-head attention (MHA)~\citep{vaswani2017attention}, we separate within-block offsets from the number of selected blocks between a key and its query: keys retain reusable within-block rotations, while queries accumulate the rotations associated with selected blocks. This yields fused attention kernels that preserve tensor-core matrix multiplication without materializing per-pair rotated keys.
For shared-KV attention (e.g., MQA~\citep{shazeer2019multiquery} or MLA~\citep{deepseekai2024deepseekv2}) with shared selections across heads, we derive a strategy that reuses key rotations across those heads.

Our experiments use the block-sparse implementation to retain compatibility with OLMo-3's MHA architecture.
After 8K mid-training, \acronym retains broadly competitive general capabilities and already shows gains in length extrapolation (Figure~\ref{fig:training-ablation}).
After extension to 64K, it improves over sparse RoPE with YaRN~\citep{peng2024yarn} at the training length by 6.14 points ($74.10 \to 80.24$) on RULER \citep{hsieh2024ruler} and 4.40 points ($42.97 \to 47.37$) on HELMET \citep{yen2024helmet} (Table~\ref{tab:stage3}).
The RULER advantage persists beyond the training context (Figure~\ref{fig:teaser}(b)): without further training or changes to positional-encoding settings, \acronym leads all evaluated variants at 128K, 256K, and 512K.
At 512K---eight times the training length---it scores 72.17, retaining approximately 90\% of its 64K score, versus 15.21 for sparse RoPE with YaRN and 70.63 for NoPE~\citep{kazemnejad2023impactnope}, the strongest extrapolation baseline (Table~\ref{tab:stage3-extrapolation}).
Controlled ablations further show that training with \acronym outperforms switching to it only at inference, with a 21.24-point RULER advantage at 256K, supporting the importance of learning the re-indexed geometry during training.

Our contributions are threefold:
\begin{itemize}[leftmargin=1.2em,itemsep=2pt,parsep=0pt]
  \item \textbf{Native co-design of position encoding and sparse attention.} To our knowledge, we are the first to co-design position encoding with sparse attention by making position reindexing \emph{native} to training. The encoding-agnostic rule preserves selected-token order, bounds relative distances by the attention budget, and recovers ordinary indexing when no tokens are skipped.
  \item \textbf{A trainable RoPE instantiation.} We develop \acronym and derive rotation-reuse strategies for block-sparse MHA and shared-KV attention with shared selections. Our fused block-sparse implementation supports parallel training with tensor-core matrix multiplication, without materializing per-pair rotated keys.
  \item \textbf{Controlled evidence for native training.} Through OLMo-3-7B continued pretraining, we demonstrate that \acronym improves long-context performance over RoPE-based baselines and extrapolates to eight times the training length without additional positional scaling. Ablations isolate the benefit of training with the reindexed positions over inference-only adaptation.
\end{itemize}

%% file: sec/2_relatedwork.tex
\section{Related Work}
\label{sec:related}

\paragraph{Position Encoding.}
Since the sinusoidal absolute encodings of \citet{vaswani2017attention}, much of the work on position encoding has aimed at making positions easier to extend to longer sequences.
Although dropping the explicit encoding has been reported to maintain the in-length performance and improve out-of-length generalization \citep{kazemnejad2023impactnope,gelberg2026drope}, Rotary Position Encoding (RoPE) \citep{su2024roformer} is still what most frontier open-weight models~\citep{liu2025deepseekv32-dsa,xu2026deepseekv4-dsa,glm53,qiu2026qwen38next-qsa,team2026gemma4} ship with---so it is where we start.
RoPE gives every token a integer position id and turns that id into a rotation of the token's query and key, using a fixed set of frequencies.
Length extension methods for RoPE-based LLMs can be thus divided into two lines: those that keep the id and reschedule the frequencies, and those that change the id itself.

The first line keeps the id that enumeration assigns and changes only the schedule that turns it into a rotation: positional interpolation \citep{chen2023posinterpolation} linearly interpolates all the frequencies according to the target length, YaRN \citep{peng2024yarn} applies an NTK-aware correction to the base frequency, LongRoPE \citep{ding2024longrope} and LongRoPE2 \citep{shang2025longrope2} search for a separate factor per frequency band, and Context-Aware RoPE \citep{veisi2025carope} lets the context modulate the frequencies. 
These methods, however, are typically tailored to a target length: the scaling factors depend on the intended extension and may need adjustment when the target changes; rescaling can affect even the short distances the model was trained on, so performance inside the original window can drop \citep{liu2024scalinglawsrope,ding2024longrope}; and coverage up to the fitted length does not by itself establish length generalization beyond it.

By contrast, the second line changes the position indices themselves, and with them what a distance measures.
As a training-time data augmentation, \citet{ruoss2023randomized} sample random, order-preserving ids so that training covers a wider range of gaps than enumeration provides, and then revert to enumeration at inference. Such a strategy incurs training-inference inconsistency, while it enlarges the context window that the model can use.
As an inference-time adaptation, StreamingLLM \citep{xiao2024streamingllm} and InfiniteHiP \citep{lee2025infinitehip} apply position reindexing on top of a model trained with contiguous ids: during generation, a subset of historical tokens are retained and assigned consecutive position ids.
Such a post-hoc repair enables generation past the training window, but it leaves the model on a position assignment it was never trained on---gaps between consecutive ids are now set by the eviction policy rather than by the text---so the model at best tolerates the re-indexing rather than exploiting it.
The mismatch compounds: ids drift from step to step as tokens are evicted, and which tokens survive is fixed by a heuristic that was never optimized jointly with the indexing.
We therefore propose \emph{native} position reindexing, an encoding-agnostic rule used during both training and inference with sparse attention.
The selector chooses which tokens to attend to; our reindexing rule assigns the selected tokens consecutive indices in their original order.
We instantiate this rule with RoPE as \acronym (see \cref{sec:dyrope}).
Other contextual-position methods construct coordinates differently: RePo \citep{li2026repo} learns a flexible position mapping that must itself generalize to longer contexts, while CoPE \citep{golovneva2024cope} adds content-dependent query--key gate accumulation to attention.
Given the sparse selection, our reindexing rule instead obtains distinct, ordered indices and budget-bounded distances by construction, using simple integer arithmetic on selected-token ranks rather than learned position prediction or gate accumulation.

\paragraph{Sparse Attention.}
Early sparse patterns were fixed by design: strided and dilated patterns in Sparse Transformer \citep{child2019sparsetransformer}, sliding windows with global tokens in Longformer \citep{beltagy2020longformer}, the window-plus-random-plus-global combination of Big Bird \citep{zaheer2020bigbird}. These patterns are static with respect to the content.

For LLM inference, dynamic sparsity that depends on the query has become the norm. Training-free methods keep a small subset of tokens in the cache, selected by accumulated attention scores \citep{zhang2023h2o,ribar2023sparq,xiao2024quest}, by off-line classification of attention heads and patterns \citep{jiang2024minference,xiao2024duoattention}, or by retrieval from an external context memory \citep{xiao2024infllm}.

More recently, sparsity has moved into training.
Pioneering explorations include BiFormer \citep{zhu2023biformer}, NSA \citep{yuan2025nsa}, and MoBA \citep{lu2026moba}, all train the model to adapt to heuristic content-aware sparsity from scratch. 
Following this line, a trend is to also optimize the sparse pattern in an end-to-end, fully differentiable manner \citep{hu2026hils,li2026sas,huang2026dashattention}.
For frontier open-weight models, however, efforts to date have taken a more incremental route: distillation-based continued pretraining (CPT) rather than training from scratch, with the selector trained to match the full-attention score distribution---the route taken by DeepSeek Sparse Attention (DSA) \citep{liu2025deepseekv32-dsa, xu2026deepseekv4-dsa}, MiniMax Sparse Attention (MSA) \citep{lai2026msa}, and the Qwen Sparse Attention (QSA) \citep{qiu2026qwen38next-qsa}.
What these share with the training-free line is that the position encoding is inherited unchanged from full attention: the tokens that survive selection keep the absolute indices they had in the original sequence, so the index range an attention head must handle still grows with the input length.
Native position reindexing is orthogonal and complementary to these sparse attention methods---it changes which indices the selected tokens carry, not which tokens are selected---and can in principle be layered on any selector above.

\paragraph{Long Context Modeling.}
Progress has been measured by both synthetic and realistic benchmarks: RULER \citep{hsieh2024ruler} probes retrieval over controlled lengths with synthetic tasks, while LongBench \citep{bai2024longbench} and HELMET \citep{yen2024helmet} target realistic ones; \citet{liu2024lostinthemiddle} document the positional bias that makes such evaluations sensitive to where the evidence sits.

Beyond the position encoding, length generalization has been attacked through the attention computation itself. SSMax \citep{nakanishi2025ssmax} and stick-breaking attention \citep{tan2025stickbreakingattn} change the normalizer so that attention does not degenerate as the number of keys grows; \citet{golowich2025role} give evidence that sparsity itself can help. On the data side, continued pretraining on longer sequences remains the main recipe for context scaling \citep{xiong2023effectivelongcontext}, at a cost that grows with the long tail of the length distribution.

Our work sits at the intersection of the three lines above. We train models with consecutive indices assigned to each query's selected tokens in their original order, bounding relative distances by the sparse budget. This lets the context grow without expanding the positional range. Unlike inference-time cache re-indexing, this geometry is used during both training and inference.
To our knowledge, this is the first work to co-design position encoding with sparse attention by making position reindexing \emph{native} to training.

%% file: sec/3_method.tex
\section{Method}
\label{sec:method}

Our approach brings position reindexing into the training of natively sparse attention models.
For each query, it uses the \emph{selection map} to assign the selected keys consecutive indices in their original order.
The resulting query-specific distances compress the gaps left by unselected tokens and are bounded by the attention budget rather than the sequence length.
Training with these indices lets the model learn the position assignments induced by sparse attention.

We develop the method in three parts.
\S\ref{sec:prelim} first defines the sparse selection setup, then reviews RoPE and formalises parallel position reindexing for a general selection map.
\S\ref{sec:dyrope} instantiates this rule as \acronym and derives structure-aware algorithms that preserve tensor-core computation through rotation reuse across queries in block-sparse MHA or across heads sharing KV and selection.
Finally, \S\ref{sec:loggate} describes the block-sparse model used in our experiments, with a learned block selector trained end to end through a count-corrected LogGate scheme \citep{li2026sas}.
This sparse backbone is held fixed across positional-encoding comparisons.

\subsection{Preliminaries: RoPE and Position Reindexing}
\label{sec:prelim}

\paragraph{Sparse attention setup.}
For a sequence of length $L$, a \emph{selection map} $\gA$ assigns each query $i$ a set of keys $\gA(i)\subseteq\{0,\dots,i\}$, with $i\in\gA(i)$ and $|\gA(i)|\le B\ll L$, where $B$ is the per-query attention budget, i.e., the maximum number of keys each query may attend to.
Position reindexing takes this map as input and assigns consecutive indices to the selected keys in their original order.
This indexing rule is independent of both the selection mechanism and the positional encoding; we use RoPE as a concrete instantiation below.

\paragraph{RoPE.} Let $\vx_0,\dots,\vx_{L-1}$ be the input and let $\vq_i,\vk_j \in \R^{D}$ be the query of token $i$ and the key of token $j$ in one attention head.
RoPE \citep{su2024roformer} encodes position as a rotation applied to the query and the key before the score is taken.
For an integer position $p$, the rotation is block-diagonal, with one $2\times2$ block per channel pair:
\begin{equation}
R(p) = \mathrm{diag}\big(R_0(p), \dots, R_{D/2-1}(p)\big),
\qquad
R_d(p) =
\begin{pmatrix}
\cos p\theta_d & -\sin p\theta_d \\
\sin p\theta_d & \cos p\theta_d
\end{pmatrix},
\qquad
\theta_d = \Theta^{-2d/D},
\label{eq:rope-block}
\end{equation}
where $\Theta$ is a fixed, predefined base and $d = 0, \dots, D/2-1$; the $d$-th block rotates channels $2d$ and $2d+1$.
Applying RoPE to the query and key gives the pre-softmax attention score $a_{ij}$:
\begin{equation}
\tilde{\vq}_i = R(i)\vq_i,
\qquad
\tilde{\vk}_j = R(j)\vk_j,
\qquad
a_{ij} = \frac{\langle \tilde{\vq}_i, \tilde{\vk}_j \rangle}{\sqrt{D}}
       = \frac{\vq_i^{\top} R(j-i) \vk_j}{\sqrt{D}},
\label{eq:rope}
\end{equation}
where the last equality holds because $R$ is orthogonal and a homomorphism, $R(p)^{\top}R(p') = R(p'-p)$.
Two structural properties of RoPE matter below: the first says \emph{what} may be changed, the second what it \emph{costs}.
\underline{(P1) Relative.} Only the distance $j-i$ enters the score, so re-indexing the pair, $(i,j) \mapsto (i',j')$, 
can edit nothing but the distance $j'-i'$.
The same fact lets the two indices be shifted by a common amount without changing any score---a freedom \S\ref{sec:dyrope} spends.
\underline{(P2) Local and inexpensive.} RoPE rotates each pair of channels independently, requiring $O(D)$ arithmetic for a $D$-dimensional vector.
These rotations can be performed within the attention kernel, avoiding global-memory writes and reads of intermediate results.

\begin{figure}[tbp]
  \centering
  \includegraphics[width=\linewidth]{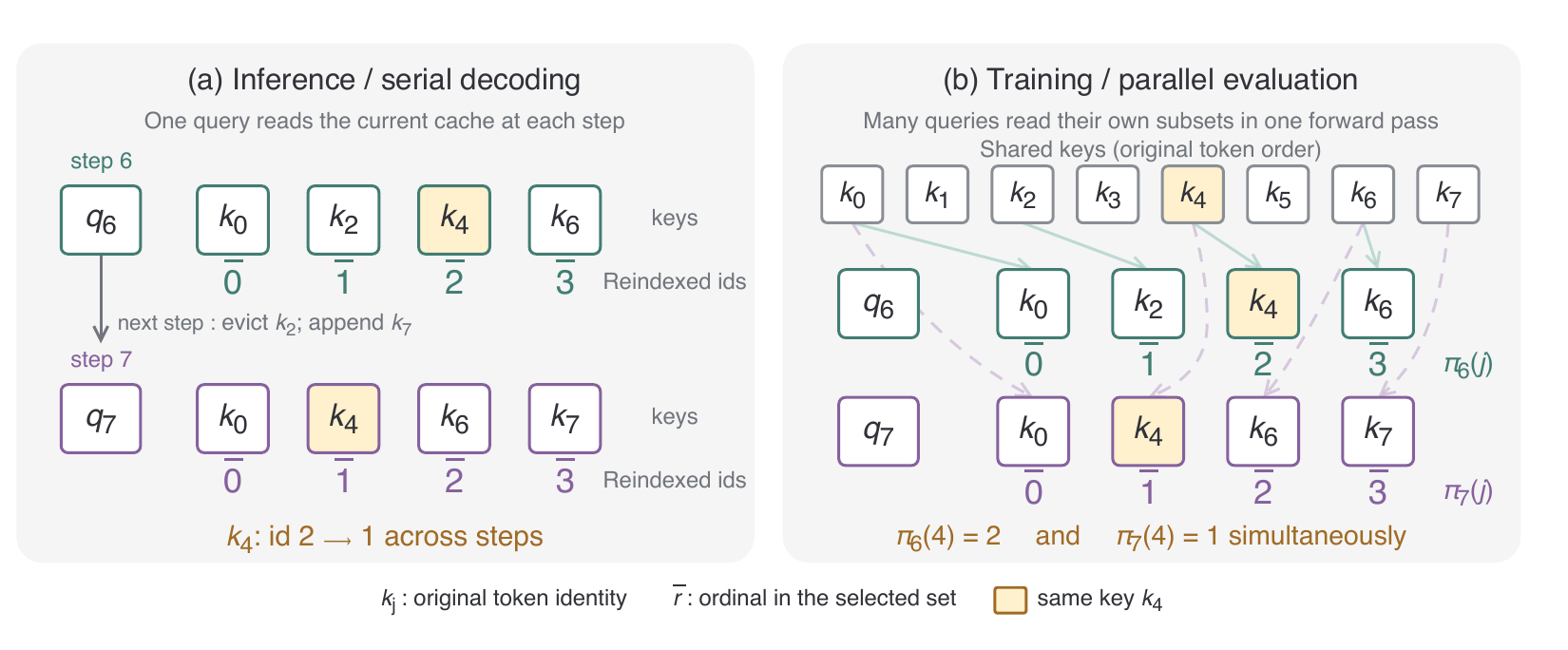}
  \caption{\textbf{Position reindexing: serial versus parallel.}
    \textbf{(a)} A cached key has one id per decoding step, which may change after eviction.
    \textbf{(b)} The same key can have different query-specific ids simultaneously: $\pi_6(4)=2$ and $\pi_7(4)=1$.
    Yellow highlights $k_4$; overbars denote reassigned ids.
    Query-specific rows are logical views, not KV copies.
  }
  \label{fig:dypi-serial-parallel}
\end{figure}

\paragraph{Position reindexing.}
Position reindexing was introduced as an inference-time repair \citep{xiao2024streamingllm,lee2025infinitehip} for streaming inference: the tokens surviving in the KV cache eviction are re-indexed by their \emph{ordinal} rather than by their original position, 
so the indices used for the attention score (\eqref{eq:rope}) stay bounded by the number of retained tokens however long the stream is.
Decoding appends one token at a time and re-indexes the cache at each step: the ids of the cached keys move, but every key still has a single id---only one query is active per sequence at each decoding step and reads the cached keys.
Training instead evaluates all $L$ queries at once, each attending to its own selected set $\gA(i)$.
A key is then read by many queries at once, each through a different subset, so it no longer has a single id: the index has to be stated per query, as illustrated in \figref{fig:dypi-serial-parallel} and formalised in Definition~\ref{def:dypi}.

\begin{definition}[Position reindexing, in parallel form]
\label{def:dypi}
Given a selection map $\gA$, the reassigned position index of key $j \in \gA(i)$ as seen by query $i$ is its ordinal in that query's active set,
\begin{equation}
\pi_i(j) = \sum_{j'=0}^{j} \mathbb I[j'\in\gA(i)] - 1 \;\in\; \{0,\dots,|\gA(i)|-1\},
\label{eq:dypi}
\end{equation}
where $\mathbb I[\cdot]$ equals 1 if its condition holds and 0 otherwise.
In particular, $\pi_i(i)=|\gA(i)|-1$.
The relative distance seen by the pair is
\begin{equation}
\delta_{ij} \;=\; \pi_i(i) - \pi_i(j)
\;=\; \sum_{j'=j+1}^{i} \mathbb I[j'\in\gA(i)]
\;\in\; \{0,\dots,|\gA(i)|-1\} \;\subseteq\; \{0,\dots,B-1\}.
\label{eq:delta}
\end{equation}
\end{definition}

\eqref{eq:dypi} is the same trick, generalised from one \emph{numbering} for a single decoding query to one numbering per query, and it refers to no encoding or sparsity pattern in particular;
on RoPE it is exactly the re-indexing (P1) permits, replacing $j-i$ by $-\delta_{ij}$, so reindexing redefines distance and nothing else.
Three properties follow directly.
\underline{(a) Dense-limit identity.} If nothing is dropped, $\gA(i) = \{0,\dots,i\}$, then $\pi_i(j) = j$: the rule degenerates exactly to ordinary indexing, hence is a strict generalisation of it and costs nothing in the dense regime.
\underline{(b) Monotone skip-compression.} $\pi_i$ is increasing in $j$, so order and causality are untouched, and what is contracted is exactly the gaps attention skips---window near, retrieved blocks in the middle, sinks \citep{xiao2024streamingllm} at the far end, all inside $[0,B-1]$ whatever $L$ is.
\underline{(c) Query-dependent.} The same key carries different indices under different queries in the same forward pass, which is what makes a per-pair encoding look unavoidable; whether it is, and what it costs, is \S\ref{sec:dyrope}.
The bounded range is the point: with contiguous indices it is the corpus that decides how much supervision each distance $d$ gets, and that tails off as $d$ grows---a $64$k-token document supplies $\approx 64$k pairs at distance $1$ but exactly one at distance $64\mathrm{k}-1$; with \eqref{eq:delta} each query contributes one pair at every distance in $[0,|\gA(i)|)$, and there is no distance at or beyond $B$.

\subsection{\acronym: RoPE with Native Position Reindexing}
\label{sec:dyrope}

Instantiating RoPE with the indices of Definition~\ref{def:dypi} gives \acronym.
We train with it rather than apply it only at inference, allowing the model to learn the re-indexed geometry (\S\ref{sec:related}).
The query and its selected keys are rotated by their ordinals in the query's active set:
\begin{equation}
\tilde{\vq}_i = R\big(\pi_i(i)\big)\vq_i,
\qquad
\tilde{\vk}_j^{\,(i)} = R\big(\pi_i(j)\big)\vk_j,
\qquad
a_{ij}
= \frac{\vq_i^{\top} R\big(\pi_i(j) - \pi_i(i)\big)\vk_j}{\sqrt{D}}
= \frac{\vq_i^{\top} R(-\delta_{ij})\vk_j}{\sqrt{D}} .
\label{eq:dyrope}
\end{equation}
The score thus depends on the position indices only through $\delta_{ij} \in \{0,\dots,B-1\}$ alone: \dyrope is a relative positional encoding whose entire domain has width $B$, independent of the sequence length $L$.
This simple reindexing fully specifies \acronym. Making it practical for parallel training, however, is non-trivial.
Note the superscript on $\tilde{\vk}_j^{\,(i)}$: the rotated key now depends on the query.
Consequently, a rotation can no longer be computed once per token and reused: one key now carries one rotation per query that selects it.

\begin{figure}[htbp]
\centering
\includegraphics[width=\textwidth]{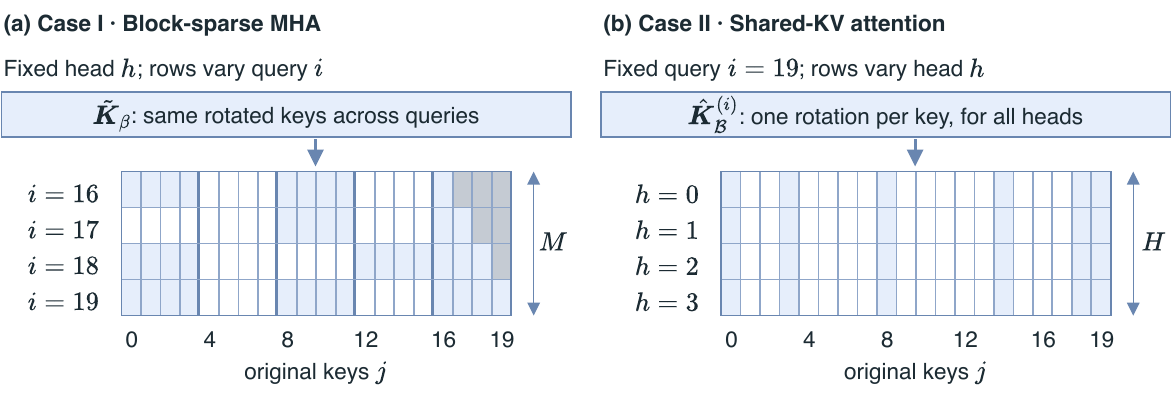}
\caption{Sparse selection and sharing constraints for \acronym (schematic).
(a) Reuse of intra-block-rotated keys across query rows is restricted to the same head in MHA and requires whole-block selection ($b=4$); different queries may select different blocks.
(b) Reuse of query-specific rotated keys across head rows is restricted to the same query token and requires those heads to share both KV and the same tokenwise selection; different queries may use different selections and rotations.
Blue denotes selected pairs; gray denotes causally masked future keys.
Appendix~\ref{app:dyrope-tiling} gives the detailed tilings.}
\label{fig:dyrope-sparse-patterns}
\end{figure}

A direct implementation, which materialises $B$ rotated active keys for all $L$ queries, is accordingly expensive: 
it costs $O(LBD)$ memory plus a second $O(LBD)$ round trip to high bandwidth memory (HBM).
Appendix~\ref{app:ref} lists a reference implementation from which this accounting is immediate.
Fusing the rotation saves memory, but each query still needs its own rotated keys.
Without additional structure, scores must be computed pair by pair with slow vector operations rather than a fast tensor-core matmul.
Kernel fusion alone therefore cannot resolve the computational inefficiency of general per-pair \acronym.
Under structural constraints common in sparse attention, however, we can design algorithms that reorganise the rotations onto individual matmul operands, preserving dense tensor-core computation; combining these structure-aware algorithms with kernel fusion makes \acronym practical.
Two common settings provide the required structure through different routes: \emph{block-sparse MHA} through the selection map (Case~I), and \emph{shared-KV attention} (MLA, MQA) with shared selections through the head axis of the matmul (Case~II).
GQA lies between these settings and combines the two strategies.
Figure~\ref{fig:dyrope-sparse-patterns} contrasts the selection supports and sharing constraints of the two settings.

\input{sec/algs/dyrope_block_mha}

\textbf{Case I: block-sparse MHA---reuse keys across queries.}
Block-sparse attention partitions the key sequence into contiguous blocks and selects whole blocks rather than individual tokens, as in NSA \citep{yuan2025nsa} and MoBA \citep{lu2026moba}.
Different queries may select different blocks, but each attends to all causally visible keys within its selected blocks.
\acronym removes gaps between selected blocks but preserves token spacing within each block.
This separates the positional rotation into an \emph{intra-block} part, determined by the token's offset within its block, and an \emph{inter-block} part, determined by how many selected blocks lie closer to the query.
The intra-block key rotation is independent of the query's selection; the inter-block part can be applied entirely to the query, allowing rotated keys to be reused.

Formally, for fixed block size $b$, token $p$ belongs to block $\beta_p=\lfloor p/b\rfloor$ with offset $p\bmod b$.
Let $\gB(i)$ be the blocks selected by query $i$, including its own block $\beta_i$, causally truncated at $i$.
For a selected block $\beta$, let $n_{i,\beta}=|\{\gamma\in\gB(i):\gamma>\beta\}|$ count the selected blocks nearer to $i$.
For $j\in\gA(i)$ in block $\beta$, $\delta_{ij}$ counts selected tokens after $j$ up to and including $i$, by \eqref{eq:delta}.
For a historical block, we count $b-1-(j\bmod b)$ tokens in the key block, $n_{i,\beta}-1$ full intervening selected blocks, and $(i\bmod b)+1$ in the query block, giving
\begin{equation}
\delta_{ij}
=\underbrace{(i\bmod b)}_{\text{query offset}}
+\underbrace{b\,n_{i,\beta}}_{\substack{\text{selected-block}\\\text{distance}}}
-\underbrace{(j\bmod b)}_{\text{key offset}}.
\label{eq:rank}
\end{equation}
In the query's own block, $n_{i,\beta_i}=0$, so the distance is just the difference of intra-block offsets.
Moving to the next farther selected block increases $n_{i,\beta}$ by one, adding $b$ regardless of how many unselected blocks are skipped.
Using RoPE's rotation composition, the score factorises as
\begin{equation}
\begin{aligned}
\tilde{\vk}_j &= R(j\bmod b)\vk_j,\qquad
\hat{\vq}_i^{(\beta)} = R(b)^{n_{i,\beta}}R(i\bmod b)\vq_i, \qquad
a_{ij} &= \frac{\big\langle\hat{\vq}_i^{(\beta)},\tilde{\vk}_j\big\rangle}{\sqrt{D}}.
\end{aligned}
\label{eq:dyrope-affine}
\end{equation}
The key carries only its fixed intra-block rotation; the query combines its intra-block rotation with one inter-block rotation $R(b)$ per nearer selected block.
All keys in block $\beta$ share the same query rotation, so their scores remain an ordinary matmul.
Algorithm~\ref{alg:dyrope-block} implements \eqref{eq:dyrope-affine} in a fused kernel.

\textbf{Case II: shared-KV attention---reuse rotations across heads.}
In MQA or MLA, suppose the $H$ heads of query token $i$ share the same selected keys.
Each key then has the same distance $\delta_{ij}$ for every head.
We can therefore \emph{rotate each key once and reuse the rotated key across all $H$ heads}.
The matmul rows are the $H$ query vectors of a single token, rather than query vectors from different tokens.
For MQA, writing $\vq_h$ for head $h$'s query at token $i$, the score is
\begin{equation}
\hat{\vk}_j^{(i)}=R(-\delta_{ij})\vk_j,
\qquad
 a_{ij}^{(h)}=\frac{\langle\vq_h,\hat{\vk}_j^{(i)}\rangle}{\sqrt{D}}.
\label{eq:dyrope-sharedkv}
\end{equation}
The cache holds unrotated keys and values; query-specific key rotations are fused into the kernel before the matmul.
For MLA~\citep{deepseekai2024deepseekv2} that uses partial RoPE, the same rotation reuse applies to the shared rotary-key component, leaving the latent-content contribution unchanged.
Keys need not be contiguous: visiting them from near to far makes $\delta_{ij}$ a running count of selected keys already visited (Algorithm~\ref{alg:dyrope-sharedkv}).
Unlike Case~I, this route supports arbitrary tokenwise selection, provided enough heads share it to amortise the rotation.

\input{sec/algs/dyrope_shared_kv}

\textbf{Discussion.}
\acronym changes position indexing, not token selection, and adds no parameters.
It is defined for any selector, but efficient execution requires reusable blocks (Case~I) or enough heads sharing KV and a selection (Case~II).
GQA lies between MHA and MQA and supports efficient block-sparse execution with \emph{sequence-major, head-minor tiling}: each tile batches query tokens, grouping each token's KV-sharing heads in adjacent rows.
Small groups provide less head-wise amortisation; without block structure or head sharing, computation remains per-pair and thus impractical.
\S\ref{sec:experiments} reports measured overhead.

\acronym is an instantiation of native position reindexing with the RoPE encoding.
However, the reindexing rule specifies position indices, not the encoding that uses them, and is therefore not tied to RoPE.
We choose RoPE for its predominant use in current LLMs and because continued pretraining of OLMo-3 enables fully transparent experiments with clear attribution and manageable compute costs.
Other encodings remain future work (see \S\ref{sec:limitation}).

\subsection{End-to-End Sparse Attention via LogGate}
\label{sec:loggate}

Native position reindexing is designed for sparse attention, whereas the released OLMo-3 checkpoint is dense.
We therefore convert it to block-sparse attention during continued pretraining, retaining its MHA architecture and enabling the efficient Case~I implementation in \S\ref{sec:dyrope}.

We use the lightweight indexer design of MiniMax Sparse Attention \citep{lai2026msa}.
Learned query/key projections produce index vectors of dimension $d_{\mathrm{idx}}$; their scaled dot products are max-pooled within each KV block to obtain its score.
Suppressing head indices,
\begin{equation}
s_{ic}=\max_{j\in\gA_{\mathrm{block}}(c),\,j\le i}
\frac{\langle\vq_i^{\mathrm{idx}},\vk_j^{\mathrm{idx}}\rangle}{\sqrt{d_{\mathrm{idx}}}}.
\label{eq:indexer-block-score}
\end{equation}
We first trained the indexer and inherited backbone with MSA's two-stage distillation recipe \citep{lai2026msa}, but did not reach full-attention parity (Appendix~\ref{app:msa-conversion}).
Keeping the indexer, we instead use a modified LogGate scheme \citep{li2026sas}, obtaining a competitive sparse baseline for integrating \dyrope.

Let $\mathcal I_i$ contain the top-scoring historical blocks, up to $k$ blocks of size $b$, and let $\kappa_i=|\mathcal I_i|$.
Together with the causally visible self-block that the $i_{th}$ query resides in, they define
\begin{equation}
\gA(i) \;=\; \gA_{\mathrm{self}}(i) \;\cup\!\! \bigcup_{c\in\mathcal I_i}\!\! \gA_{\mathrm{block}}(c),
\qquad
|\gA(i)| \;\le\; B \;=\; (k+1)b .
\label{eq:active}
\end{equation}
For $\kappa_i>0$, LogGate injects the selected block scores into attention as additive log-biases:
\begin{equation}
\begin{aligned}
\eta_{ic}&=\begin{cases}
\displaystyle
\underbrace{s_{ic}-\log\sum_{d\in\mathcal I_i}\exp(s_{id})}_{\text{log-softmax}}
+\underbrace{\log\kappa_i}_{\substack{\text{block-count}\\\text{correction}}},
& c\in\mathcal I_i,\\[1ex]
0, & c=c(i),
\end{cases}\\[1ex]
p_{ij}&=\frac{\exp(a_{ij}+\eta_{i,c(j)})}
{\sum_{v\in\gA(i)}\exp(a_{iv}+\eta_{i,c(v)})},
\qquad j\in\gA(i).
\end{aligned}
\label{eq:loggate-main}
\end{equation}
Here $c(j)$ identifies key $j$'s block, so $c(i)$ is the self-block; when $\kappa_i=0$, only the self-block is used.
Top-$k$ indices remain discrete, but the language-modeling loss trains the indexer through the selected log-gates.
Our $+\log\kappa_i$ correction restores unit historical gates when selected scores are equal, removing count-induced self-block bias (Appendix~\ref{app:loggate-correction}).

All components share a fixed KV-block grid, with the query's own block causally truncated, so \acronym uses Algorithm~\ref{alg:dyrope-block}.
Our positional-encoding comparisons hold the LogGate scheme, attention budget $B$, and training recipe fixed.

%% file: sec/algs/dyrope_block_mha.tex
\begin{algorithm}[t]
\small
\caption{
\dyrope fused into a FlashAttention-style block-sparse MHA kernel (Case~I), one head shown.
\dyropeswatch{} marks what \dyrope changes relative to standard RoPE: 
The intra-block rotations use token offsets: $R(j\bmod b)$ for cached keys and $R(i\bmod b)$ for the initial query.
Walking blocks from near to far, the inter-block rotation updates only the query by $R(b)$ after each selected block; skipped blocks cause no update.
Figure~\ref{fig:dyrope-block-tiling} in Appendix~\ref{app:dyrope-tiling} illustrates the sparse pattern and query-major tiling.
}
\label{alg:dyrope-block}
\begin{algorithmic}[1]
\State{\textbf{pre-pass} (separate elementwise kernel)}
\ForAll{token $p$ and channel pair $(2r,2r{+}1)$}                        \Comment{\text{\textbf{grid-parallel}; one thread per element}}
  \State{$\tilde\vk_p \gets R(p \dyropestep{\bmod b})\,\vk_p$}            \Comment{\text{\dyropecomment{intra-block} rotation; written into the KV cache once, reused by every query}}
  \State{$\hat\vq_p \gets R(p \dyropestep{\bmod b})\,\vq_p$}              \Comment{\text{\dyropecomment{intra-block} rotation; the position the query has in its own block}}
\EndFor
\State{}
\State{\textbf{main kernel} (fused block-sparse multi-head attention)}
\ForAll{CTA indexed by (batch, head, query tile $\gI$ of $M$ rows)}      \Comment{\text{\textbf{grid-parallel}; one CTA per tile, no sharing}}
  \State{load $\hat\mQ$, $\mV$ base pointers; $\vo_i\gets\vzero$, $\ell_i\gets 0$, $\mu_i\gets-\infty$ for $i\in\gI$}
  \State{$\gB\gets\bigcup_{i\in\gI}\gB(i)$, restricted to causally visible blocks, decreasing block id}   \Comment{\text{tile-level union, precomputed}}
  \State{$\vm_\beta\gets$ bitmask of $\beta$ over the $M$ rows}          \Comment{\text{precomputed once per tile; drives both the mask and the rotation}}
  \For{each block $\beta\in\gB$}                                         \Comment{\text{\textbf{sequential}: carries $(\mu_i,\ell_i,\vo_i)$ and the rotation}}
    \State{issue async loads of $\tilde\mK_\beta,\mV_\beta$ into shared memory}  \Comment{\text{overlaps the MMA of the previous block}}
    \State{$\mA\gets\hat\mQ\,\tilde\mK_\beta^{\top}/\sqrt{D}$}           \Comment{\text{MMA on tensor cores; \emph{all} $M$ rows, warp-parallel}}
    \State{$\mA[i,j]\gets-\infty$ where $\vm_\beta[i]=0$, and where $\beta$ is diagonal and $j>i$}  \Comment{\text{elementwise, warp-parallel}}
    \State{$\mu_i^{\mathrm{new}}\gets\max\bigl(\mu_i,\ \max_j \mA[i,j]\bigr), \forall i \in \gI$}  \Comment{\text{row-parallel}}
    \State{$\mP\gets\exp\bigl(\mA-\mu_i^{\mathrm{new}}\vone^{\top}\bigr)$;
           $\ell_i\gets \ell_i\,e^{\mu_i-\mu_i^{\mathrm{new}}}+\sum_j\mP[i,j]$}
    \State{$\vo_i\gets \vo_i\,e^{\mu_i-\mu_i^{\mathrm{new}}}+\mP[i,:]\,\mV_\beta$;
           $\mu_i\gets\mu_i^{\mathrm{new}}$}                             \Comment{\text{accumulate on tensor cores}}
    \State{\dyropestep{$\hat\vq_i\gets R(b)\,\hat\vq_i$ if $\vm_\beta[i]=1, \forall i \in \gI$}} \Comment{\text{\dyropecomment{inter-block rotation; vector units, in the shadow of the next MMA}}}
  \EndFor
  \State{$\vo_i\gets\vo_i/\ell_i$; write $\vo_i$ and $L_i=\mu_i+\log\ell_i, \forall i \in \gI$}   \Comment{\text{row-parallel epilogue}}
\EndFor
\end{algorithmic}
\end{algorithm}

%% file: sec/algs/dyrope_shared_kv.tex
\begin{algorithm}[t]
\small
\caption{\dyrope in shared-KV attention, illustrated in MQA form (Case~II), for one query token $i$; all $H$ heads share its selected keys and values.
\dyropeswatch{} marks the changes to standard RoPE: cache unrotated keys, then rotate each selected key once inside the kernel and reuse it across heads.
Keys are visited nearest-first.
Figure~\ref{fig:dyrope-sharedkv-tiling} in Appendix~\ref{app:dyrope-tiling} illustrates the sparse pattern and head-major tiling.}
\label{alg:dyrope-sharedkv}
\begin{algorithmic}[1]
\State{\textbf{KV cache layout} (what the cache holds)}
\ForAll{token $p$}                                                    \Comment{\text{\textbf{grid-parallel}; one thread per element}}
  \State{store \dyropestep{unrotated} key $\vk_p$ and value $\vv_p$}          \Comment{\text{\dyropecomment{key rotation is query-dependent}}}
\EndFor
\State{}
\State{\textbf{main kernel} (fused MQA with RoPE)}
\ForAll{CTA indexed by (batch, query token $i$)}                      \Comment{\text{\textbf{grid-parallel}; one CTA per query, no sharing}}
  \State{load $\mQ\in\R^{H\times D}$ with rows $\vq_h$; $\vo_h\gets\vzero$, $\ell_h\gets0$, $\mu_h\gets-\infty$ for $h=1,\dots,H$}
  \State{\dyropestep{$s\gets0$}}                                      \Comment{\text{\dyropecomment{rotation counter, carried by the tile loop; $\delta_{ii}=0$}}}
  \For{each key tile $\gB\subseteq\gA(i)$, nearest first}             \Comment{\text{\textbf{sequential}: carries $(\mu_h,\ell_h,\vo_h)$ and $s$}}
    \State{issue async loads of $\{\vk_j,\vv_j\}_{j\in\gB}$, packed near-to-far} \Comment{\text{tokenwise selection; no contiguity}}
    \ForAll{key $j$ at zero-based rank $t$ in the tile, nearest first} \Comment{\text{vector units, warp-parallel}}
      \State{\dyropestep{$\hat\vk_j^{(i)}\gets R(-s-t)\,\vk_j$}} \Comment{\text{\dyropecomment{$\delta_{ij}=s+t$: one rotation per key}}}
    \EndFor
    \State{\dyropestep{$s\gets s+|\gB|$}}                             \Comment{\text{\dyropecomment{keys consumed so far; carried to the next (farther) tile}}}
    \State{$\mA\gets\mQ\bigl(\hat\mK_\gB^{(i)}\bigr)^\top/\sqrt{D}$} \Comment{\text{single MMA, $M=H$ rows}}
    \State{$\mu_h^{\mathrm{new}}\gets\max\bigl(\mu_h,\ \max_j\mA[h,j]\bigr)$}  \Comment{\text{row-parallel}}
    \State{$\mP\gets\exp\bigl(\mA-\mu_h^{\mathrm{new}}\vone^\top\bigr)$;
           $\ell_h\gets\ell_h\,e^{\mu_h-\mu_h^{\mathrm{new}}}+\sum_j\mP[h,j]$}
    \State{$\vo_h\gets\vo_h\,e^{\mu_h-\mu_h^{\mathrm{new}}}+\mP[h,:]\,\mV_\gB$;
           $\mu_h\gets\mu_h^{\mathrm{new}}$}                          \Comment{\text{values shared by all $H$ rows}}
  \EndFor
  \State{$\vo_h\gets\vo_h/\ell_h$; write $\vo_h$ and $L_h=\mu_h+\log\ell_h$}   \Comment{\text{row-parallel epilogue}}
\EndFor
\end{algorithmic}
\end{algorithm}

%% file: sec/4_experiments.tex
\section{Experiments}
\label{sec:experiments}

We evaluate \acronym through matched continued-pretraining experiments on OLMo-3-7B, asking three questions:
(1) Can it improve long-context performance while preserving general capabilities?
(2) How well does it extrapolate beyond the training context without further training or positional retuning?
(3) What does native training offer beyond inference-only reindexing?
We also examine training dynamics and measure the incremental forward-kernel cost of inter-block rotation.

\subsection{Experimental Setup}
\label{sec:experimental-setup}

\paragraph{Training and evaluation setup.}
We initialize all models from the publicly released OLMo-3-7B Stage-1 checkpoint, pretrained on approximately 5.93T tokens \citep{olmo3}.
Following the OLMo-3 recipe, we conduct Stage~2 (mid-training) on approximately 50B tokens at an 8K context length.
We halve the official Stage-2 token budget from 100B to 50B to fit mid-training within a one-week compute budget on 64 NVIDIA H20 GPUs.
Stage~3 (long-context extension) continues training from the corresponding Stage-2 checkpoints on an additional 50B tokens at 64K.
Within each stage, the models share the data mixture, training-token budget, and optimizer schedule.
We use 200 warmup steps to a peak learning rate of $2.07\times10^{-4}$, followed by linear decay.
We follow the OLMo-3 base-model evaluation setup for math, general, and code benchmarks, and evaluate long-context performance on RULER and HELMET.

\paragraph{Model variants.}
We compare Full-RoPE, our baseline retaining full attention in OLMo-3's global layers and RoPE with no architectural modifications, with four sparse variants.
All models retain OLMo-3's 32-layer MHA architecture, including 24 sliding-window layers with a 4,096-token window.
Full-RoPE retains dense attention in the remaining eight global layers, while the sparse variants use the same LogGate-based block-sparse architecture (\S\ref{sec:loggate}) and attention budget in these layers.
\textbf{(1) Sparse-RoPE} applies standard RoPE using the original token positions in both the main attention and the indexer.
\textbf{(2) Sparse-HoPE} \citep{chen2025hope} retains the higher-frequency components of RoPE and sets lower-frequency components to zero, with the cutoff determined by the training context length; this encoding is applied to both the main attention and the indexer.
\textbf{(3) Sparse-NoPE} \citep{kazemnejad2023impactnope,gelberg2026drope} removes explicit positional encoding from both components.
\textbf{(4) Sparse-\acronym} assigns consecutive indices to each query's selected tokens in their original order and applies RoPE using these query-specific indices, while its indexer uses NoPE.
In Stage~3, Full-RoPE and Sparse-RoPE apply YaRN \citep{peng2024yarn} with a scaling factor of 8 over the 8K context to their global layers, to adapt these RoPE baselines to the eightfold context extension from 8K to 64K.
Sparse-HoPE and Sparse-\acronym use no positional scaling. All sliding-window layers retain RoPE with unscaled frequencies because their 4,096-token window, and hence their relative-distance range, remains unchanged.
Full training, implementation, and evaluation details are provided in Appendix~\ref{app:experimental-details}.

\subsection{Results}
\label{sec:experimental-results}

\input{sec/tables/stage2_8k_in_length_eval}

\paragraph{General capability after mid-training (8K-token context window).}
Table~\ref{tab:stage2} reports the results after 50B tokens of 8K mid-training.
Sparse-\acronym matches or exceeds Full-RoPE and Sparse-RoPE on most general benchmarks, obtaining the highest scores on MMLU, MMLU-Pro, ARC-C, and WinoGrande, while trailing Full-RoPE on BBH and code generation.
It also performs competitively on math, with the highest scores on GSM8K and MATH-500.
These results show that native training with position reindexing preserves broad task capability under sparse attention.
The advantage of \acronym becomes clearer at longer contexts: at RULER 4K, all five models score between 92.52 and 93.74 (4K fits within one sliding window), but at 8K Sparse-\acronym reaches 85.17, exceeding Full-RoPE and Sparse-RoPE by 4.82 and 3.14, respectively.
Thus, the benefit of \acronym is already visible during mid-training: it preserves general capability while improving retrieval at the upper end of the training context.

\input{sec/tables/stage3_64k_long_eval}

\paragraph{Long-context performance after extension (64K-token context window).}
Table~\ref{tab:stage3} reports the results after 50B tokens of 64K long-context extension.
Sparse-\acronym stays within 0.16 of Full-RoPE on RULER at 4K and within 0.08 at 8K, and achieves the highest scores at 16K, 32K, and 64K.
Its margin over Full-RoPE grows from 2.61 at 16K to 4.36 at 32K and 8.29 at 64K.
At the 64K training limit, Sparse-\acronym scores 80.24, improving on Sparse-RoPE's 74.10 by 6.14.
The advantage of \acronym thus becomes more pronounced as retrieval spans a longer context.
The gains also extend to HELMET's broader long-context tasks.
At 64K, Sparse-\acronym scores 47.37, exceeding Full-RoPE and Sparse-RoPE by 4.05 and 4.40, respectively.
These improvements are obtained with YaRN applied to both RoPE baselines, while \acronym uses no additional positional scaling.
Together, the Stage-2 and Stage-3 results show that \acronym preserves broad task capability through mid-training and develops stronger long-context performance through context extension.

\input{sec/tables/stage3_64k_extrapolation_eval}

\paragraph{Length extrapolation beyond the training context.}
Table~\ref{tab:stage3-extrapolation} evaluates the Stage-3 models beyond their 64K training context, without further training or changes to their positional-encoding settings.
Sparse-\acronym achieves the highest RULER scores among all five models at both 128K and 256K, reaching 77.53 and 75.19, respectively.
At 256K, both RoPE baselines degrade sharply (Full-RoPE to 5.58, Sparse-RoPE to 23.30), leaving Sparse-\acronym ahead by 69.61 and 51.89, respectively; it also exceeds Sparse-HoPE by 15.41.
Sparse-NoPE is the strongest sparse baseline at these lengths; \acronym's advantage over it grows from 2.35 at 128K to 2.75 at 256K.
At 512K, eight times the training context, Sparse-\acronym still scores 72.17, retaining approximately 90\% of its 64K score.
HELMET likewise shows strong retention: Sparse-\acronym scores 46.61 at 128K, only 0.76 below its 64K result, while Full-RoPE and Sparse-RoPE drop to 23.45 and 30.41.
On the five-category HELMET extension at 256K (excluding RAG and reranking; Appendix~\ref{app:experimental-details}), Sparse-\acronym scores 51.13, exceeding Sparse-NoPE by 5.14 points.
Together, the RULER and HELMET results show that \acronym's long-context gains extend beyond the training context.

\begin{figure}[htbp]
  \centering
  \includegraphics[width=\linewidth]{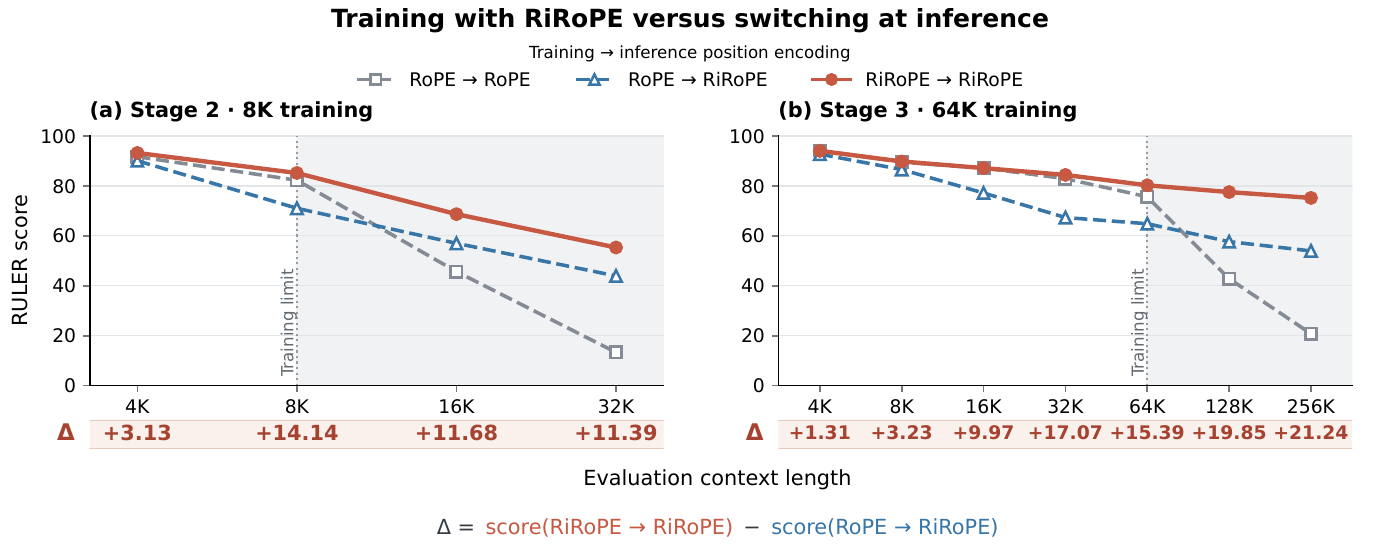}
  \caption{
    \textbf{Training with \acronym versus switching at inference.}
    All configurations use a NoPE indexer; arrows in the legend denote training and inference encodings.
    The $\Delta$ row reports trained \acronym minus inference-only switching at each length.
    Shading marks evaluation beyond the training length.
    RoPE uses YaRN in Stage~3.
  }
  \label{fig:training-ablation}
\end{figure}

\paragraph{Benefits of native training for position reindexing.}
Figure~\ref{fig:training-ablation} compares using position reindexing throughout continued pretraining (RiRoPE) with introducing it only at inference in a RoPE-trained model, without updating its weights.
All configurations use a NoPE indexer and follow the same training recipe.
At the Stage-2 training length of 8K, the model trained with \acronym reaches a RULER score of 85.17, exceeding the inference-only switch's 71.03 by 14.14.
The benefit of training therefore already appears within the context length covered during training.
At 16K and 32K, inference-only switching improves the original RoPE model's extrapolation performance, while training with \acronym delivers further gains of 11.68 and 11.39, respectively.
The same pattern persists after the Stage-3 extension to 64K: the model trained with \acronym scores 80.24 at the training length, exceeding inference-only switching by 15.39.
The margin grows to 19.85 and 21.24 at 128K and 256K, respectively, showing that the benefits of training extend to longer contexts.
These results strongly support incorporating position reindexing during training rather than introducing it only at inference: it allows the model to learn the positional relationships induced by sparse selection, which benefits both in-range and extrapolation performance.

\begin{figure}[t]
  \centering
  \includegraphics[width=\linewidth]{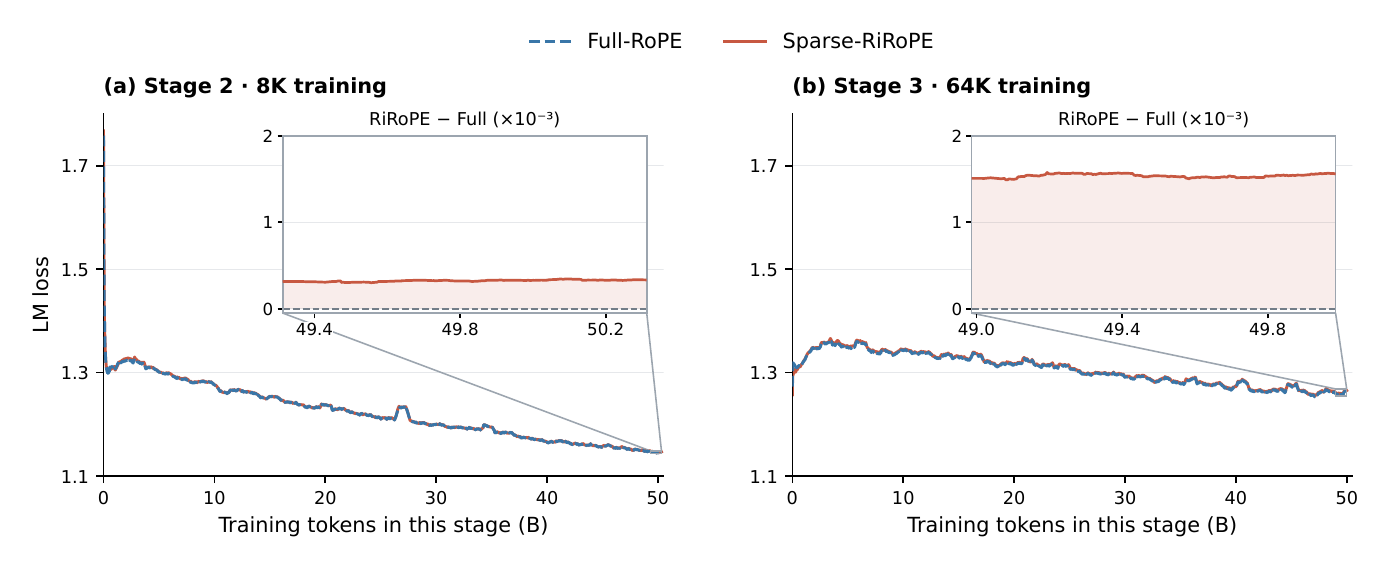}
  \caption{
    \textbf{LM loss during continued pretraining.}
    Full-RoPE and Sparse-\acronym follow the Stage-2 (8K, left) and Stage-3 (64K, right) recipes.
    Curves show trailing moving averages over 1B training tokens.
    Insets show the LM-loss gap (\acronym minus Full-RoPE) over the boxed final 1B-token regions on a shared vertical scale; the dashed zero line denotes equal loss.
    Full-RoPE uses YaRN in Stage~3.
  }
  \label{fig:lm-loss}
\end{figure}

\paragraph{Training loss dynamics.}
Figure~\ref{fig:lm-loss} compares the LM-loss trajectories of Full-RoPE and Sparse-\acronym during 8K mid-training and 64K long-context extension.
Sparse-\acronym closely tracks Full-RoPE across both stages.
The insets zoom in on the final 1B tokens of each stage, revealing slightly higher LM loss for Sparse-\acronym than for Full-RoPE.
Over the final 5B training tokens of each stage, the mean loss gap is $3.37\times10^{-4}$ in Stage~2 and $1.52\times10^{-3}$ in Stage~3.
Despite this residual loss gap, Sparse-\acronym outperforms Full-RoPE on RULER and HELMET at 64K and retains stronger performance beyond the training context (Tables~\ref{tab:stage3} and~\ref{tab:stage3-extrapolation}).
Thus, slightly higher LM loss can coexist with stronger long-context task performance and length extrapolation, consistent with prior findings that LM loss alone does not fully capture downstream performance \citep{qiu2026qwen38next-qsa}.

\begin{figure}[t]
\centering
\includegraphics[width=\textwidth]{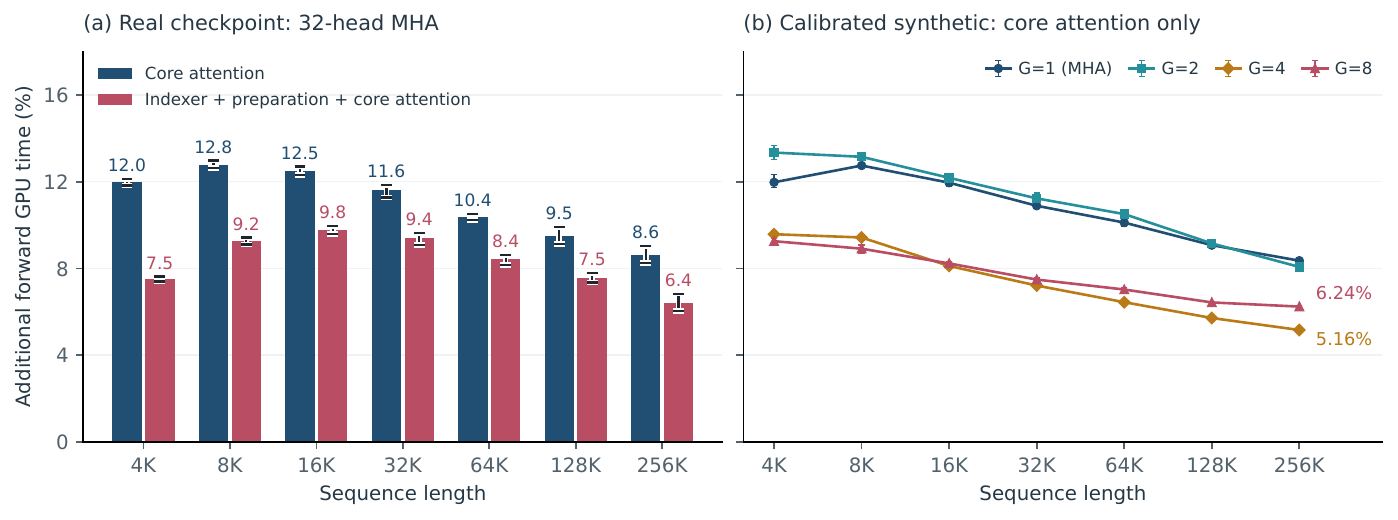}
\caption{Forward overhead of incremental inter-block rotation on NVIDIA H20: the overhead decreases with context length and is lower for GQA with group sizes $G{=}4$ and $G{=}8$ than for MHA.
(a)~Real OLMo-3 MHA workloads at 4K--256K, relative to core attention or indexer+preparation+core attention baselines, where preparation comprises gate processing, block-union construction, and output initialization.
(b)~Calibrated synthetic workloads with 32 query heads and $G{=}H_q/H_{kv}\in\{1,2,4,8\}$; each configuration uses its own PE-off baseline.
Error bars: 95\% bootstrap CIs. Q/K pre-rotation excluded; lengths above 64K exceed training.}
\label{fig:dyrope-efficiency}
\end{figure}

\paragraph{Overhead of \acronym.}
The additional cost of \acronym comes from online inter-block query rotation within the sparse attention kernel (Algorithm~\ref{alg:dyrope-block}).
Figure~\ref{fig:dyrope-efficiency} reports this forward overhead on an NVIDIA H20.
Our kernel provides a unified implementation of GQA and MHA.
The group size $G=H_q/H_{kv}$ is the number of query heads sharing each key/value head, with MHA corresponding to the special case $G=1$.
The overhead decreases with context length: across 4K--256K, the MHA percentage falls from 12.75\% to 8.36\%, and on real OLMo-3 workloads at 256K it is 8.62\% of core attention time and 6.42\% when including indexer and preparation costs.
GQA group sharing further reduces the cost: with 32 query heads, increasing $G$ from 1 (MHA) to 8 lowers the overhead to 6.24\%--9.25\%, and both $G{=}4$ and $G{=}8$ remain below MHA at every length.
The added core-attention overhead is thus already modest, even with a kernel that has not yet been fully tuned.
At longer contexts, the more important scaling concern is the indexer: with a fixed selection budget, the useful core-attention work per query is bounded independently of context length, whereas indexer scoring grows with the visible history.
This makes the indexer the asymptotic bottleneck, consistent with the long-context profiling results reported in IndexCache \citep{bai2026indexcache}.
Absolute increments and measurement details are in Appendix~\ref{app:efficiency-measurement}.

%% file: sec/tables/stage2_8k_in_length_eval.tex
\begin{table}[t]
\centering
\caption{
  \textbf{Stage-2 (8K mid-training) results.}
  All models start from the same OLMo-3-7B pre-training checkpoint and are trained on 50B tokens of 8K mid-training data.
  They share OLMo-3's architecture and differ only in the 8 global-attention layers (dense or block-sparse) and their position encoding; ours is shaded.
  Sparse attention uses the LogGate scheme described in Section~\ref{sec:loggate}.
  Avg.\ is the mean over all 15 benchmarks.
  Best per row in \tbest{bold}, second \tsecond{underlined}.
}
\label{tab:stage2}
\vspace{4pt}
{\tablesize
\setlength{\tabcolsep}{4.5pt}%
\renewcommand{\arraystretch}{1.18}%
\begin{NiceTabular}{l *{5}{w{c}{1.5cm}}}
\CodeBefore
  \rectanglecolor{oursbg}{4-6}{22-6}
  \rowcolor{groupbg}{3,7,15,18}
\Body
\toprule
 & \textbf{Full Attn.} & \multicolumn{4}{c}{\textbf{Sparse Attn.}} \\
\cmidrule(lr){2-2} \cmidrule(lr){3-6}
\textbf{Benchmark} & RoPE & RoPE & HoPE~\citep{chen2025hope} & NoPE~\citep{kazemnejad2023impactnope} & \tours{\dyrope} \\
\midrule
\multicolumn{6}{l}{\textbf{Math}} \\
GSM8K               & 73.35 & 74.48 & \tsecond{74.58} & \tsecond{74.58} & \tbest{74.83} \\
Minerva MATH        & 35.99 & \tsecond{36.96} & \tbest{37.04} & 36.04 & 36.52 \\
MATH-500            & 36.77 & 38.60 & \tsecond{38.73} & 38.24 & \tbest{39.56} \\
\multicolumn{6}{l}{\textbf{General}} \\
MMLU                & 65.48 & \tsecond{65.76} & 65.72 & 65.67 & \tbest{65.86} \\
MMLU-Pro            & \tsecond{34.80} & 34.70 & 34.75 & 34.23 & \tbest{35.21} \\
BBH                 & \tbest{64.80} & \tsecond{63.75} & 63.54 & 63.39 & 61.66 \\
HellaSwag           & 77.57 & \tsecond{77.67} & 77.63 & 77.41 & \tbest{77.76} \\
ARC-C               & \tsecond{84.32} & 84.02 & 83.67 & 83.47 & \tbest{84.59} \\
WinoGrande          & \tsecond{85.84} & 85.74 & 85.68 & 85.47 & \tbest{86.49} \\
DROP                & 60.40 & 63.30 & \tsecond{64.33} & \tbest{64.89} & 62.35 \\
\multicolumn{6}{l}{\textbf{Code}} \\
HumanEval           & \tbest{45.33} & 41.29 & \tsecond{43.46} & 42.93 & 42.21 \\
MBPP                & \tbest{49.01} & \tsecond{48.62} & 47.84 & 45.68 & 48.06 \\
\multicolumn{6}{l}{\textbf{Long context}} \\
RULER 4K            & 92.65 & \tbest{93.74} & 92.52 & 93.02 & \tsecond{93.20} \\
RULER 8K            & 80.35 & 82.03 & 81.04 & \tsecond{83.53} & \tbest{85.17} \\
HELMET 8K           & \tbest{46.47} & \tsecond{45.93} & 43.43 & 44.63 & 44.99 \\
\midrule
Avg.          & 62.21 & \tsecond{62.44} & 62.26 & 62.21 & \tbest{62.56} \\
\bottomrule
\end{NiceTabular}
}
\end{table}

%% file: sec/tables/stage3_64k_long_eval.tex
\begin{table}[t]
\centering
\caption{
  \textbf{Stage-3 (64K long-context extension) results on RULER and HELMET.}
  Each model continues from its own Stage-2 checkpoint (Table~\ref{tab:stage2}) on 50B tokens of 64K long-context data.
  The RoPE models are extended with YaRN (factor 8 over the 8K Stage-2 length), whereas HoPE and \acronym (ours, shaded) use no scaling.
  Best per column in \tbest{bold}, second \tsecond{underlined}.
}
\label{tab:stage3}
\vspace{4pt}
{\tablesize
\setlength{\tabcolsep}{4pt}%
\renewcommand{\arraystretch}{1.18}%
\begin{NiceTabular}{l l *{5}{w{c}{1.0cm}} w{c}{1pt} *{4}{w{c}{1.0cm}}}
\CodeBefore
  \rectanglecolor{oursbg}{7-2}{7-12}
\Body
\toprule
 & & \multicolumn{5}{c}{\textbf{RULER}} & & \multicolumn{4}{c}{\textbf{HELMET}} \\
\cmidrule(lr){3-7} \cmidrule(lr){9-12}
\textbf{Attention} & \textbf{Position} & 4K & 8K & 16K & 32K & 64K & & 8K & 16K & 32K & 64K \\
\midrule
Full Attn. & RoPE + YaRN~\citep{peng2024yarn} & 93.87 & \tsecond{89.86} & 84.53 & 80.04 & 71.95 & & 47.98 & 47.10 & 44.75 & 43.32 \\
\arrayrulecolor{black!30}\midrule\arrayrulecolor{black}
\Block[l]{4-1}{\shortstack[l]{Sparse Attn.}}
  & RoPE + YaRN~\citep{peng2024yarn} & 93.61 & 87.75 & 83.07 & 80.70 & 74.10 & & \tbest{48.56} & 47.28 & \tsecond{46.78} & 42.97 \\
  & HoPE~\citep{chen2025hope} & \tbest{94.90} & \tbest{89.92} & 83.45 & 77.36 & 71.37 & & 46.10 & 44.90 & 45.14 & 43.88 \\
  & NoPE~\citep{kazemnejad2023impactnope} & \tsecond{94.16} & 89.37 & \tsecond{86.28} & \tsecond{83.39} & \tsecond{78.86} & & 47.81 & \tbest{48.07} & \tbest{48.13} & \tbest{49.20} \\
  & \tours{\acronym} & 94.03 & 89.78 & \tbest{87.14} & \tbest{84.40} & \tbest{80.24} & & \tsecond{48.09} & \tsecond{47.41} & 46.35 & \tsecond{47.37} \\
\bottomrule
\end{NiceTabular}
}
\end{table}

%% file: sec/tables/stage3_64k_extrapolation_eval.tex
\begin{table}[t]
\centering
\caption{
  \textbf{Length extrapolation beyond the 64K training length.}
  The same Stage-3 models as Table~\ref{tab:stage3}, evaluated at context lengths above the 64K they were trained on.
  The \tref{grey} 64K columns provide in-range reference scores and are excluded from ranking. Best per extrapolation column in \tbest{bold}, second \tsecond{underlined}.
  $^\dagger$Five-category HELMET extension (Appendix~\ref{app:experimental-details}).
}
\label{tab:stage3-extrapolation}
\vspace{4pt}
{\tablesize
\setlength{\tabcolsep}{4pt}%
\renewcommand{\arraystretch}{1.18}%
\begin{NiceTabular}{l l w{c}{1.0cm} w{c}{1pt} *{3}{w{c}{1.0cm}} w{c}{1pt} w{c}{1.0cm} w{c}{1pt} *{2}{w{c}{1.0cm}}}
\CodeBefore
  \rectanglecolor{oursbg}{7-2}{7-12}
\Body
\toprule
 & & \multicolumn{5}{c}{\textbf{RULER}} & & \multicolumn{4}{c}{\textbf{HELMET}} \\
\cmidrule(lr){3-7} \cmidrule(lr){9-12}
\textbf{Attention} & \textbf{Position} & 64K & & 128K & 256K & 512K & & 64K & & 128K & 256K$^\dagger$ \\
\midrule
Full Attn. & RoPE + YaRN~\citep{peng2024yarn} & \tref{71.95} & & 13.06 & \phantom{0}5.58 & \phantom{0}2.81 & & \tref{43.32} & & 23.45 & 17.30 \\
\arrayrulecolor{black!30}\midrule\arrayrulecolor{black}
\Block[l]{4-1}{\shortstack[l]{Sparse Attn.}}
  & RoPE + YaRN~\citep{peng2024yarn} & \tref{74.10} & & 42.23 & 23.30 & 15.21 & & \tref{42.97} & & 30.41 & 28.07 \\
  & HoPE~\citep{chen2025hope} & \tref{71.37} & & 68.04 & 59.78 & 56.93 & & \tref{43.88} & & 42.38 & 39.50 \\
  & NoPE~\citep{kazemnejad2023impactnope} & \tref{78.86} & & \tsecond{75.18} & \tsecond{72.44} & \tsecond{70.63} & & \tref{49.20} & & \tsecond{46.45} & \tsecond{45.99} \\
  & \tours{\acronym} & \tref{80.24} & & \tbest{77.53} & \tbest{75.19} & \tbest{72.17} & & \tref{47.37} & & \tbest{46.61} & \tbest{51.13} \\
\bottomrule
\end{NiceTabular}
}
\end{table}

%% file: sec/5_limitation.tex
\section{Limitation and Future Work}
\label{sec:limitation}

\paragraph{Beyond RoPE.}
Our reindexing rule specifies how positions are indexed, independently of the encoding that consumes those indices.
Our experiments instantiate it only with RoPE as \acronym, preserving compatibility with OLMo-3's pretrained weights and continued-pretraining (CPT) pipeline.
Future work could explore more position encoding schemes, such as ALiBi~\citep{press2021alibi} and learned relative-position mechanisms, including Inkling's relative attention~\citep{burtenshaw2026inkling} and Swin Transformer's relative position bias~\citep{liu2021swin}.
Their effectiveness under native position reindexing remains untested in this work.

\paragraph{MLA/MQA/GQA models.}
Although our method describes efficient implementation strategies for \acronym with MLA~\citep{deepseekai2024deepseekv2} and MQA~\citep{shazeer2019multiquery} through cross-head reuse when heads share selected keys, 
our model experiments are limited to OLMo-3's MHA architecture.
Compared with OLMo-3, the MLA models we considered, such as those in the DeepSeek~\citep{deepseekai2024deepseekv2} and Kimi~\citep{kimiteam2025kimik2} families, are much larger and provide open weights without the accompanying pretraining data and complete training pipelines, making similarly controlled, affordable experiments difficult.
End-to-end validation of both efficiency and downstream performance on MLA/MQA and GQA~\citep{ainslie2023gqa} models remains future work.

\paragraph{Training from scratch.}
We validate \acronym through CPT rather than pretraining from scratch.
Starting from a capable pretrained model enables meaningful downstream evaluation within a controlled compute budget.
From-scratch studies face both higher training costs and evaluation challenges: reducing the model scale and relying on language-modeling loss alone may obscure downstream differences~\citep{mistralai2026mistrallarge4}.
From-scratch training may differ in dynamics from CPT~\citep{gelberg2026drope}, so generalization to that setting requires further validation.

\paragraph{Modeling skipped gaps.}
Native position reindexing encodes relative order and distance within each query's selected token set, but does not explicitly represent the gaps skipped in the original sequence.
Consequently, adjacent selected tokens receive adjacent position indices even when they were far apart before selection.
An open question is whether auxiliary gap information offers benefits beyond selected-token geometry, and how to incorporate it without sacrificing compact indexing.

\paragraph{Causal video generation.}
Causal video DiTs such as CausVid~\citep{yin2024causvid} and Self Forcing~\citep{huang2025selfforcing} provide a natural setting for extending \dyrope beyond language models.
The goal is to train on short clips yet generate longer videos in real time under a fixed active-context budget.
Key questions are whether to apply position reindexing only along the temporal axis or across all spatiotemporal axes, and how to preserve spatial structure and efficient KV reuse.
Long-horizon quality, temporal consistency, and latency require validation, since positional extrapolation alone does not resolve autoregressive error accumulation.

%% file: sec/6_conclusion.tex
\section{Conclusion}
\label{sec:conclusion}

We introduced native position reindexing, co-designing position assignment with sparse attention by making position reindexing native to training.
Consecutive indexing of selected tokens preserves their order while bounding relative distances by the attention budget rather than sequence length.
Our RoPE instantiation, \acronym, supports parallel training through rotation reuse and fused kernels that preserve tensor-core matrix multiplication.
Controlled OLMo-3-7B continued-pretraining experiments demonstrate improved long-context performance and length extrapolation: at 512K, eight times the training length, \dyrope retains approximately 90\% of its 64K RULER score without further training or positional retuning.
Ablations show that inference-only reindexing does not recover the benefits of training with reindexed positions.
Together, these findings show that sparse attention should shape not only which tokens interact, but also the relative distances used in those interactions.

%% file: sec/acknowledgments.tex
\section*{Acknowledgments}
We thank Xiang Hu and Xinyu Wei for integrating \dyrope into HiLS \citep{hu2026hils} and testing the integration after we had completed the initial validation of \dyrope and before we aligned our training setup with the official OLMo3 training pipeline. Their experiments further strengthened our confidence in pursuing this work.

%% file: sec/7_appendix.tex
\section{Reference Implementation of \acronym}
\label{app:ref}

The following PyTorch reference implements \eqref{eq:dyrope} for an arbitrary causal selection map $\gA$, without assuming block sparsity or sharing selections across queries.
For one attention head, \texttt{q} and \texttt{k} contain unrotated queries and keys, and each row of \texttt{active} lists the distinct tokens in $\gA(i)$ in increasing original position, with $i$ as its last valid entry.
Here $B$ bounds the number of active \emph{tokens}, not blocks; \texttt{valid} marks padding, which is excluded from the position count.

The cumulative valid count assigns each selected key its ordinal $\pi_i(j)$, and the row's total count gives $\pi_i(i)=|\gA(i)|-1$.
We keep the query unrotated and rotate each gathered key by $R(\pi_i(j)-\pi_i(i))=R(-\delta_{ij})$, which yields the same score as rotating both operands by their respective ordinals.
The function returns these pre-softmax scores only; the rotation uses adjacent channel pairs and the frequency base $\Theta$ of \eqref{eq:rope-block}.

\begingroup
\tcbset{title={Direct RiRoPE reference for arbitrary sparse selection},colbacktitle=codeframe,coltitle=black,fonttitle=\small\sffamily\bfseries,breakable}
\newcommand{\dyropememorymark}{\bfseries\color[rgb]{0.00,0.42,0.45}}
\lstset{
  emph={k_sel,k_rot},
  emphstyle=\dyropememorymark,
  morecomment=[l][\dyropememorymark]{\# MEMORY:}
}
\begin{pycodebox}
import math
import torch

def rope(x, pos, theta):
    """R(pos) on adjacent channel pairs; theta is the RoPE base."""
    D = x.shape[-1]
    assert D %
    freq = theta ** (-torch.arange(
        0, D, 2, device=x.device, dtype=torch.float32) / D)
    angle = pos.float().unsqueeze(-1) * freq
    cos, sin = angle.cos(), angle.sin()
    x0, x1 = x.float()[..., 0::2], x.float()[..., 1::2]
    return torch.stack((x0 * cos - x1 * sin,
                        x0 * sin + x1 * cos), dim=-1).flatten(-2)

def qk_with_rirope(q, k, active, valid, theta):
    """
    q, k: [L, D], unrotated; all tensors on the same device.
    active: [L, B] int64, valid IDs strictly increasing, self last.
    valid: [L, B] bool; each row contains its query token.
    Returns FP32 scores [L, B], with padding masked to -inf.
    """
    safe_ids = active.masked_fill(~valid, 0)
    pi_key = valid.long().cumsum(-1) - 1
    pi_query = valid.sum(-1, keepdim=True) - 1

    k_sel = k[safe_ids]  # MEMORY: [L, B, D] key copies; O(LBD) storage.
    k_rot = rope(k_sel, pi_key - pi_query, theta)  # MEMORY: another O(LBD) tensor, in FP32.
    scores = (q.float().unsqueeze(1) * k_rot).sum(-1)
    scores = scores / math.sqrt(q.shape[-1])
    return scores.masked_fill(~valid, -torch.inf)
\end{pycodebox}
\endgroup

The same key can receive different rotations in different query rows because its ordinal depends on the selected set, rather than its original position.
Both \texttt{k\_sel} and \texttt{k\_rot} have shape $L\times B\times D$: explicitly forming these tensors costs $O(LBD)$ storage, and writing then reading the rotated keys adds $O(LBD)$ memory traffic, as discussed in \S\ref{sec:dyrope}.
This is the direct, general implementation that motivates the structured factorizations in that section, not one of those optimized kernels.
When $\gA(i)=\{0,\ldots,i\}$, the ordinals equal the original positions and the scores reduce to ordinary RoPE.

\section{Sparse Patterns and Tiling for \acronym}
\label{app:dyrope-tiling}

\begin{figure}[p]
\centering
\includegraphics[width=\textwidth,height=0.86\textheight,keepaspectratio]{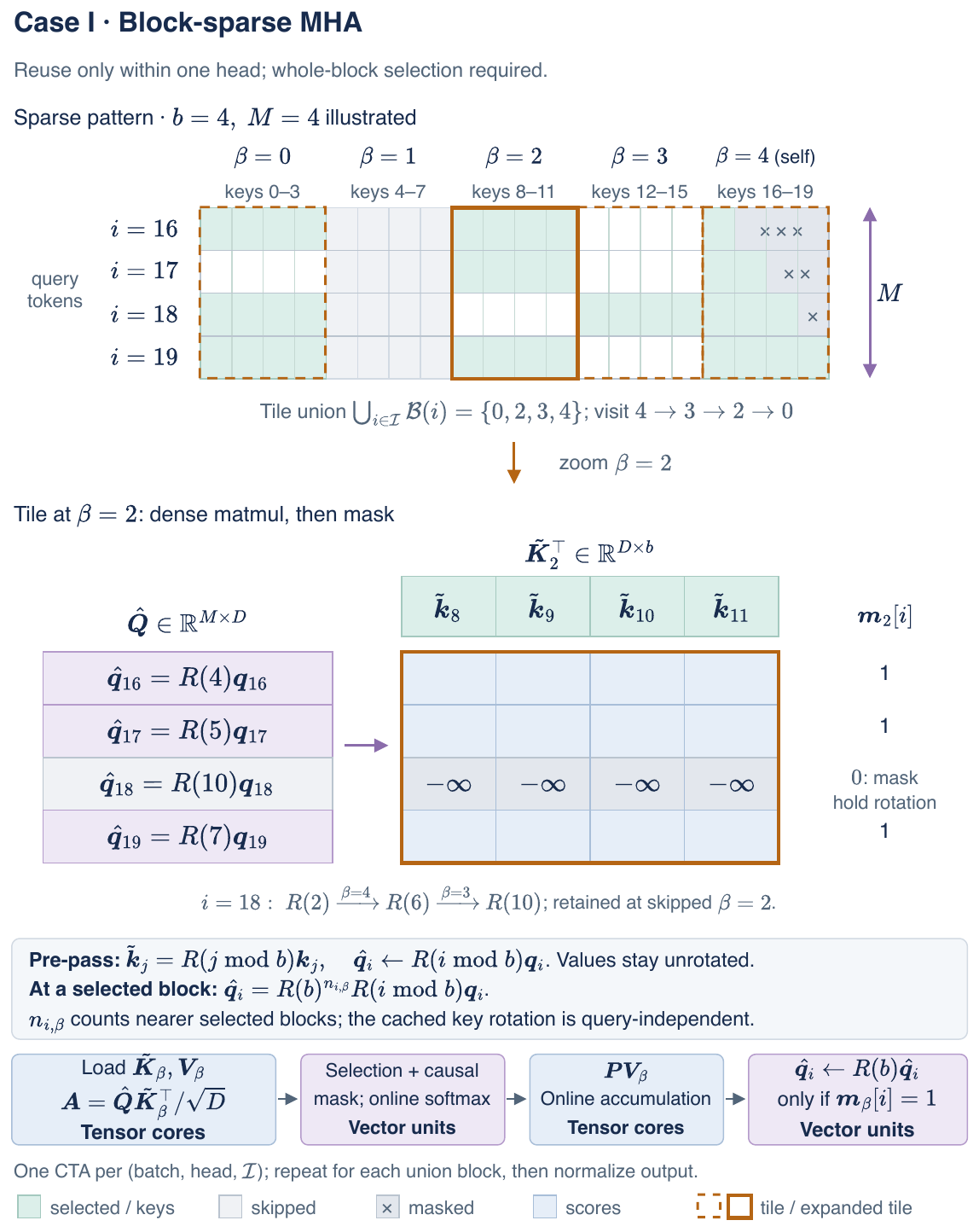}
\caption{Block-sparse MHA: query-major tiling and kernel dataflow (Case~I; Algorithm~\ref{alg:dyrope-block}).
The tile $\gI=\{16,17,18,19\}$ visits the union $\{0,2,3,4\}$ nearest-first; at $\beta=2$, row $i=18$ retains $R(10)\vq_{18}$ and is masked without a rotation update.
The lower panels show the pre-pass, cached operands, and fused computation, distinguishing tensor-core from vector-unit work.
Tile sizes are schematic.}
\label{fig:dyrope-block-tiling}
\end{figure}

\begin{figure}[p]
\centering
\includegraphics[width=\textwidth,height=0.86\textheight,keepaspectratio]{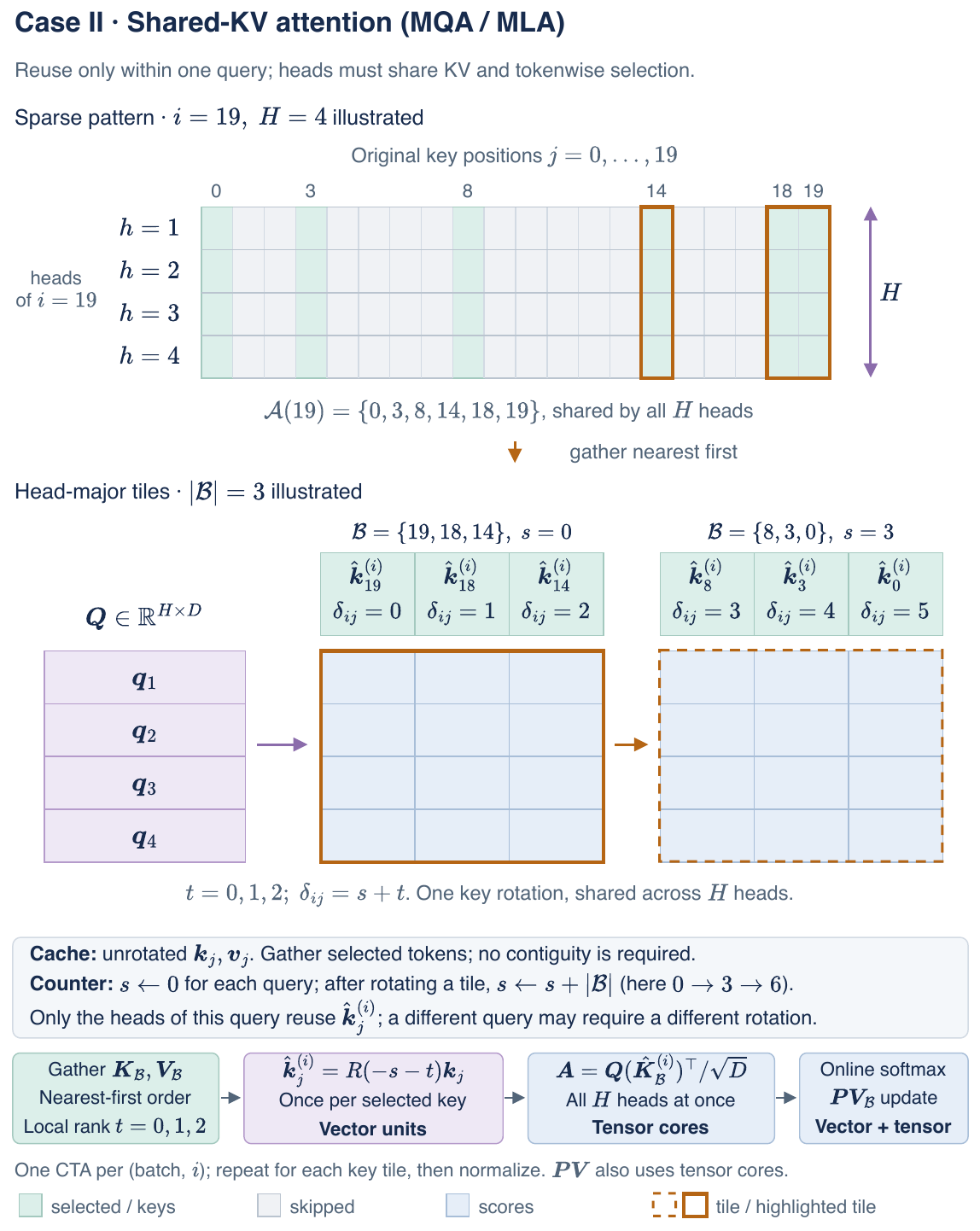}
\caption{Shared-KV attention: head-major tiling and kernel dataflow (Case~II; Algorithm~\ref{alg:dyrope-sharedkv}).
All $H$ heads of query $i=19$ share KV and $\gA(i)=\{0,3,8,14,18,19\}$; the two packed tiles use backward ranks $0$--$2$ and $3$--$5$.
The lower panels show unrotated caching, the per-query counter, and the gather--rotate--matmul--accumulate flow; each rotated key is reused across heads, not across query tokens.
Tile sizes are schematic.
}
\label{fig:dyrope-sharedkv-tiling}
\end{figure}

Figures~\ref{fig:dyrope-block-tiling} and~\ref{fig:dyrope-sharedkv-tiling} detail the sparse patterns, tensor-core tilings, and kernel dataflows of Algorithms~\ref{alg:dyrope-block} and~\ref{alg:dyrope-sharedkv}, respectively, including cache contents, rotation updates, and the constraints on reuse.
The examples use schematic tile sizes: Case~I places query tokens on the matmul row axis, whereas Case~II places the heads of one query token on that axis.

\section{LogGate Count Correction for Block-Sparse Attention}
\label{app:loggate-correction}

We derive the block-count correction in \S\ref{sec:loggate} starting from a unified indicator-based formulation of dense and sparse attention.

\subsection{A Unified Indicator Formulation of Attention}
\label{app:loggate-definition}

Suppressing layer and head indices, let $I_i(j)\in\{0,1\}$ indicate whether query $i$ attends to a causally available key $j\le i$.
Writing $a_{ij}$ for the base attention logit and $\vv_j$ for the value, attention with any nonempty selected support can be expressed as
\begin{equation}
\vo_i^{\mathrm{hard}}
=\sum_{j=0}^{i}
\frac{I_i(j)\exp(a_{ij})}
{\sum_{v=0}^{i}I_i(v)\exp(a_{iv})}\,\vv_j.
\label{eq:loggate-hard-attention}
\end{equation}
The restriction $j\le i$ enforces causality, while $I_i(j)$ specifies which of these keys participate in attention.
For dense causal attention, $I_i(j)\equiv1$, and \eqref{eq:loggate-hard-attention} reduces to the usual softmax over all causal keys.
For sparse attention with active-key set $\gA(i)$, $I_i(j)=\mathbb I[j\in\gA(i)]$, so only selected keys contribute to the numerator and denominator.
Dense and sparse attention thus share the same formula and differ only in the choice of indicator.

\subsection{Specializing to Block-Sparse Top-$k$ Attention}
\label{app:loggate-block-selection}

We now specialize this formulation to the block-sparse selection rule in \S\ref{sec:loggate}.
Let $c(j)=\lfloor j/b\rfloor$ identify the size-$b$ block containing token $j$, and let $\mathcal C_i$ contain all historical blocks preceding the query's self-block $c(i)$.
Using the indexer's block scores $s_{ic}$ from \eqref{eq:indexer-block-score}, we select
\begin{equation}
\kappa_i=\min(k,|\mathcal C_i|),
\qquad
\mathcal I_i=\operatorname{Top}_{\kappa_i}\{s_{ic}:c\in\mathcal C_i\},
\label{eq:loggate-support}
\end{equation}
where $\mathcal I_i$ is a set of historical block IDs and $k$ excludes the self-block, which is always retained separately.
The token indicator is then determined by a shared blockwise indicator:
\begin{equation}
I_i(j)=h_{i,c(j)},
\qquad
h_{ic}=\begin{cases}
1, & c=c(i)\ \text{or}\ c\in\mathcal I_i,\\
0, & \text{otherwise}.
\end{cases}
\label{eq:loggate-indicator}
\end{equation}
All causal keys in the same block share one selection decision; the self-block is still truncated by $j\le i$ in \eqref{eq:loggate-hard-attention}.
The historical block indicators therefore sum to $\kappa_i$, whereas the number of active tokens also depends on block size and the causal prefix of the self-block:
\begin{equation}
\sum_{c\in\mathcal C_i}h_{ic}=\kappa_i,
\qquad
|\gA(i)|=\kappa_i b+r_i,
\qquad r_i=(i\bmod b)+1,
\label{eq:loggate-key-count}
\end{equation}
for complete historical blocks and an unpadded sequence, with $\gA(i)$ defined in \eqref{eq:active}.
This distinction between block count and token count determines the calibration of the surrogate below.

\subsection{A Count-Preserving Surrogate Gives the LogGate Offset}
\label{app:loggate-surrogate}

The top-$k$ indicators are discrete.
To expose the selected scores to the LM objective without changing the sparse support, replace the retained historical indicators with nonnegative, score-dependent weights whose sum remains $\kappa_i$.
For $\kappa_i>0$, define
\begin{equation}
\rho_{ic}=\frac{\exp(s_{ic})}{\sum_{d\in\mathcal I_i}\exp(s_{id})},
\qquad
\gamma_{ic}=\begin{cases}
\kappa_i\rho_{ic}, & c\in\mathcal I_i,\\
1, & c=c(i),\\
0, & \text{otherwise}.
\end{cases}
\label{eq:loggate-surrogate}
\end{equation}
The historical surrogate then matches the indicator count and has unit mean on the selected blocks:
\begin{equation}
\sum_{c\in\mathcal C_i}\gamma_{ic}=\kappa_i,
\qquad
\frac{1}{\kappa_i}\sum_{c\in\mathcal I_i}\gamma_{ic}=1.
\label{eq:loggate-mean-one}
\end{equation}
For block-sparse attention, replacing $I_i(j)=h_{i,c(j)}$ with $\gamma_{i,c(j)}$ in \eqref{eq:loggate-hard-attention} yields an additive log-bias on the retained support because $\gamma_{i,c(j)}\exp(a_{ij})=\exp(a_{ij}+\log\gamma_{i,c(j)})$.
For selected historical blocks, this bias is exactly
\begin{equation}
\eta_{ic}=\log\gamma_{ic}
=s_{ic}-\log\sum_{d\in\mathcal I_i}\exp(s_{id})+\log\kappa_i,
\qquad c\in\mathcal I_i.
\label{eq:loggate-corrected}
\end{equation}
The self-block has $\eta_{i,c(i)}=0$, and dropped blocks remain masked.
Hence the resulting attention is
\begin{equation}
\vo_i=\sum_{j\in\gA(i)}p_{ij}\vv_j,
\qquad
p_{ij}=\frac{\exp(a_{ij}+\eta_{i,c(j)})}
{\sum_{v\in\gA(i)}\exp(a_{iv}+\eta_{i,c(v)})},
\label{eq:loggate-attention}
\end{equation}
recovering \eqref{eq:loggate-main}.
The offset $+\log\kappa_i$ therefore converts a probability vector, whose entries sum to one, into a surrogate for $\kappa_i$ retained-block indicators.
More generally, a shared offset $\alpha_i$ gives total historical gate weight $\exp(\alpha_i)\sum_c\rho_{ic}=\exp(\alpha_i)$, so count matching uniquely requires $\alpha_i=\log\kappa_i$.
Because the softmax is over blocks, using $\log(\kappa_i b)$ would incorrectly give mean gate $b$; using $\log k$ would also miscalibrate prefixes with fewer than $k$ historical blocks.
The normalization support matters as well: an all-candidate softmax followed by top-$k$ masking generally retains less than unit probability mass, so adding $\log\kappa_i$ alone would not give \eqref{eq:loggate-mean-one}.

Here ``surrogate'' means a differentiable reweighting on the selected support, not a differentiable replacement for top-$k$ membership.
In particular, $\gamma_{ic}$ is not a probability and may exceed one.
Requiring both $0\le\gamma_{ic}\le1$ and total weight $\kappa_i$ on exactly $\kappa_i$ retained historical blocks would force every retained gate to be one, leaving no score-dependent reweighting.
Our construction preserves nonnegativity and the indicator count, but deliberately does not impose that upper bound.

\subsection{Why Count Matching Removes the Self-Block Bias}
\label{app:loggate-bias}

For $\kappa_i>0$, fix the selected support and base logits.
Define the \emph{unnormalized attention masses} of the self-block and selected history under unshifted gating as
\begin{equation}
S_i=\sum_{j\in\gA_{\mathrm{self}}(i)}\exp(a_{ij}),
\qquad
H_i=\sum_{c\in\mathcal I_i}\rho_{ic}\sum_{j\in\gA_{\mathrm{block}}(c)}\exp(a_{ij}).
\label{eq:loggate-unnormalized-masses}
\end{equation}
They sum to the ordinary-softmax denominator $S_i+H_i$, rather than to one.
Adding $+\log\kappa_i$ to historical logits multiplies their gates by $\kappa_i$, while the self-block gate stays one.
The total historical attention weight therefore changes as
\begin{equation}
P_{i,\mathrm{hist}}^{\mathrm{unshifted}}=\frac{H_i}{S_i+H_i}
\quad\longrightarrow\quad
P_{i,\mathrm{hist}}^{\mathrm{corrected}}=\frac{\kappa_i H_i}{S_i+\kappa_i H_i}.
\label{eq:loggate-history-normalization}
\end{equation}
Unlike a common shift of all logits, this historical-only offset cannot cancel: $S_i$ is not scaled.
It increases historical attention weight when $\kappa_i>1$ and leaves it unchanged when $\kappa_i=1$.

When selected scores are equal, the unshifted historical gates are $\rho_{ic}=1/\kappa_i$, whereas ungated sparse attention on the same support---\eqref{eq:loggate-hard-attention} with the binary selection mask but no continuous LogGate reweighting---assigns every retained block a unit gate.
Unshifted gating thus suppresses historical exponential mass by a factor of $\kappa_i$ relative to this baseline, without changing the self-block mass.
The correction restores every retained gate to one and hence exactly recovers the ungated attention distribution for arbitrary base logits, including a causally truncated self-block.
For nonuniform scores, it preserves unit mean gating but does not guarantee equality to that baseline.

For the experimental \texttt{softmax1} in \eqref{eq:exp-softmax1}, replace $S_i$ by $1+S_i$ in \eqref{eq:loggate-history-normalization}; the non-cancellation argument and equal-score recovery still hold when both variants use the same normalization.

\subsection{Training and Implementation}
\label{app:loggate-implementation}

For a fixed selected support, the log-gate is differentiable in its selected scores:
\begin{equation}
\frac{\partial\eta_{ic}}{\partial s_{id}}
=\mathbb I[c=d]-\rho_{id},
\qquad c,d\in\mathcal I_i.
\label{eq:loggate-jacobian}
\end{equation}
This path trains the indexer through the LM objective without differentiating the selected block IDs.
The count offset has no score derivative on a fixed support, but it changes the attention weights and therefore can change the resulting LM gradients.

We compute \eqref{eq:loggate-corrected} with a stable log-sum-exp over the gathered historical scores, using FP32 reductions and excluding invalid slots from both the normalization and $\kappa_i$.
Each $\eta_{ic}$ is broadcast over the keys of its block and added to their logits before attention normalization; the self-block receives zero bias, and causal and selection masks remain in force.
This scalar blockwise bias changes neither the selected-token positions nor the intra-/inter-block rotations used by \dyrope.
If $\kappa_i=0$, only the self-block is used and neither the historical log-softmax nor $\log\kappa_i$ is evaluated.
If $\kappa_i=1$, the single historical gate is identically one and its gate-score derivative is zero.
The same corrected gates remain active during training and inference.

\section{Experimental Details}
\label{app:experimental-details}

This appendix specifies the Full-RoPE and LogGate-based configurations used in the main experiments.
The MSA-KL conversion protocols and results are reported separately in Appendix~\ref{app:msa-kl}.

\subsection{Training Setup}
\label{app:exp-training}

\paragraph{Initial checkpoint and data.}
All Stage-2 models inherit the backbone weights of the OLMo-3-7B checkpoint \texttt{Olmo-3-1025-7B-stage1-step1413814} and its tokenizer.
For Stage~2, we randomly partition the 100B-token Ingredient 1 version of the released Dolmino mid-training mixture\footnote{\url{https://huggingface.co/datasets/allenai/dolma3_dolmino_mix-100B-1125}} to obtain a 50B-token subset, which is shared across models and used at 8K context.
Apart from this reduced token budget and the corresponding shorter training schedule, we follow the official Stage-2 training recipe \citep{olmo3}.
Stage~3 uses 50B tokens from the Longmino long-context mixture\footnote{\url{https://huggingface.co/datasets/allenai/dolma3_longmino_mix-100B-1125}} at 64K context.
Each Stage-3 model starts from its own Stage-2 checkpoint at step 24,000.
Only model weights are transferred between stages; optimizer state, learning-rate schedule, and the stage-specific step counter are initialized anew.
We report the final checkpoints at steps 24,000 and 11,921 for Stage~2 and Stage~3, respectively.

\paragraph{Optimization.}
Table~\ref{tab:exp-training} gives the settings shared by the main comparisons within each stage.
The nominal token budget is the step count multiplied by the batch token budget; the number of tokens actually consumed is slightly lower because packed batches need not fill their budgets exactly.
The global batch sizes denote equivalent fixed-length sequences, rather than the number of source documents packed into a batch.
All reported training runs use seed 42.

\begin{table}[htbp]
\centering
\caption{\textbf{Training settings for the main experiments.} Token budgets are nominal; the same settings apply to Full-RoPE and the four sparse variants within each stage.}
\label{tab:exp-training}
{\tablelargesize
\setlength{\tabcolsep}{6pt}
\renewcommand{\arraystretch}{1.12}
\begin{tabularx}{\linewidth}{l >{\centering\arraybackslash}X >{\centering\arraybackslash}X}
\toprule
Setting & Stage 2 & Stage 3 \\
\midrule
Context length & 8,192 & 65,536 \\
Training steps & 24,000 & 11,921 \\
Token budget & 50.33B & 50.00B \\
Global batch (equivalent sequences) & 256 & 64 \\
Global batch token budget & 2,097,152 & 4,194,304 \\
Peak learning rate & $2.07\times10^{-4}$ & $2.07\times10^{-4}$ \\
Learning-rate warmup & 200 steps & 200 steps \\
Learning-rate decay & Linear & Linear \\
Optimizer & AdamW with SkipStep & AdamW with SkipStep \\
Adam $(\beta_1,\beta_2)$ / $\epsilon$ & $(0.9,0.95)$ / $10^{-8}$ & $(0.9,0.95)$ / $10^{-8}$ \\
Weight decay & 0.1 & 0.1 \\
Gradient clipping & 1.0 & 1.0 \\
Random seed & 42 & 42 \\
Parallelism / compute precision & FSDP2 / BF16 & FSDP2 / BF16 \\
\bottomrule
\end{tabularx}}
\end{table}

Embedding weights are excluded from weight decay.
We use the OLMo-3 z-loss coefficient of $10^{-5}$ during training and enable gradient checkpointing.
Following the OLMo-3 training recipe \citep{olmo3}, we use the SkipStep rule implemented in OLMo-core\footnote{\url{https://github.com/allenai/OLMo-core/blob/main/src/olmo_core/optim/skip_step_optimizer.py}}: an optimizer update is skipped when the current loss or gradient norm exceeds its recent mean by six standard deviations, using a 128-step history. Consumed-token and schedule counters still advance on that step.
The models are trained with FSDP2, without context or sequence parallelism.

\subsection{Model Variants and Initialization}
\label{app:exp-variants}

\paragraph{Shared architecture.}
The backbone has 32 layers, hidden size 4,096, and 32 query and 32 key/value heads of dimension 128.
Every fourth layer is a global-attention layer, giving eight global layers and 24 sliding-window layers with a 4,096-token window.
Full-RoPE retains the global layers; the sparse variants replace them with block-sparse MHA. All variants retain OLMo-3's RMSNorm over each full 4,096-dimensional query and key projection before splitting it into attention heads.
Within each sparse layer, four index-query heads of dimension 128 share a single index-key head.
Routing group $r\in\{0,1,2,3\}$ serves attention heads $8r,\ldots,8r+7$.
The selected block IDs and LogGate scores are shared within a routing group, while the main attention retains its independent query, key, and value heads.

\paragraph{Positional encodings.}
Table~\ref{tab:exp-variants} separates the encoding of the main attention from that of the indexer.
Model names refer to the main-attention encoding.
In the result tables, Full Attn.\ denotes Full-RoPE, and the LogGate rows or columns correspond to the named sparse variants.
All configurations preserve ordinary RoPE with $\theta=500{,}000$ in the sliding-window layers.

\begin{table}[htbp]
\centering
\caption{\textbf{Position encodings in the eight global layers.} Stage-3 scaling and the HoPE cutoff are specified in the text. Full-RoPE has no indexer.}
\label{tab:exp-variants}
{\tablelargesize
\setlength{\tabcolsep}{12pt}
\renewcommand{\arraystretch}{1.15}
\begin{tabular}{lll}
\toprule
Model & Main attention & Indexer \\
\midrule
Full-RoPE & RoPE & --- \\
Sparse-RoPE & RoPE & RoPE \\
Sparse-HoPE & HoPE & HoPE \\
Sparse-NoPE & NoPE & NoPE \\
Sparse-\dyrope & \dyrope & NoPE \\
\bottomrule
\end{tabular}}
\end{table}

For Stage~3, Full-RoPE and Sparse-RoPE use YaRN \citep{peng2024yarn} in their global layers, with scaling factor 8, original context length 8,192, $\theta=500{,}000$, $\beta_{\mathrm{fast}}=32$, and $\beta_{\mathrm{slow}}=1$.
YaRN multiplies the rotary sine and cosine tables by $m=1+0.1\ln8\approx1.20794$, scaling the resulting query--key inner products by $m^2$ relative to rotations with the same frequencies and unit amplitude.
Sparse-RoPE applies the same rotary tables to its main and indexer queries and keys.
YaRN is not applied to the sliding-window layers or to Sparse-HoPE, Sparse-NoPE, and Sparse-\dyrope.

For HoPE, the 128-dimensional rotary encoding contains 64 frequency pairs.
At training context length $L$, we keep frequencies $\omega_d\ge 4\pi/L$ and set the remaining frequencies to zero.
This retains 32 frequency pairs in Stage~2 and 42 in Stage~3, in both the main attention and the indexer.
The cutoff remains fixed during evaluation, including at lengths beyond the training context.
For each query, \dyrope assigns the selected tokens consecutive ordinal indices starting at zero, preserving their original order. It applies the ordinary RoPE frequency basis using these indices in place of the original sequence positions.

\paragraph{Indexer initialization and training.}
The main-attention and other backbone parameters are copied from the pretrained checkpoint.
For each routing group, the index-query projection is initialized by averaging the pretrained query-projection weights of its eight attention heads.
The shared index-key projection is initialized by averaging the key-projection weights across all 32 heads.
These means are computed in FP32, and the indexer's query/key RMSNorm weights are initialized to one.
After initialization, the index projections are independent trainable parameters.
The backbone and indexer are trained jointly through the LM objective and LogGate, without a separate indexer-distillation warmup or stop-gradient on the indexer's hidden-state inputs.

\paragraph{Inference-time switching control.}
The training ablation in Figure~\ref{fig:training-ablation} additionally uses a model trained with RoPE in its main attention and NoPE in its indexer.
We evaluate the same checkpoint with either RoPE or \dyrope in the main attention, without updating any weights.
Both evaluation branches retain the NoPE indexer, matching the indexer encoding of the trained Sparse-\dyrope model.

\subsection{Implementation Details}
\label{app:exp-implementation}

\paragraph{Block selection and attention.}
The indexer computes separate query and key projections, applies RMSNorm separately to each 128-dimensional index head, and scores tokens by their dot product scaled by $1/\sqrt{128}$.
The maximum score among the causally visible tokens in a block becomes the block score.
For each query and routing group, we select at most 16 historical blocks of 128 tokens and always retain the query's current block.
The latter is truncated by causality, giving an active-key bound of $B=(16+1)\times128=2{,}176$.
If fewer historical blocks are available, all are retained.

The selected historical scores receive the count-corrected LogGate transform of \S\ref{app:loggate-implementation}; the current block has zero log-bias.
Invalid slots are excluded from the normalization and count, and the score reductions use FP32.
The block choices remain discrete, while gradients through the selected scores train the indexer.
LogGate biases remain active at inference.

The main-experiment sparse layers use the \texttt{softmax1} normalization as part of our LogGate design, which adds a unit term to the attention denominator.
Writing $\eta_{i,c(j)}$ for the LogGate bias, the weights used by this implementation are
\begin{equation}
p_{ij}^{\mathrm{impl}}
=\frac{\exp\!\big(a_{ij}+\eta_{i,c(j)}\big)}
{1+\sum_{v\in\gA(i)}\exp\!\big(a_{iv}+\eta_{i,c(v)}\big)},
\qquad j\in\gA(i).
\label{eq:exp-softmax1}
\end{equation}
The extra term represents a zero-logit entry with no value contribution.
This normalization is held fixed across all four sparse positional-encoding variants.

\paragraph{Data packing.}
We tokenize plaintext online, append an end-of-sequence token to each document, and split documents longer than the stage's context length.
A mapping dataset with a seeded shuffled sampler supplies these segments to a dynamic packing loader.
Packed segments receive position IDs that restart at zero; cumulative sequence lengths identify their boundaries and prevent attention across segments.
Boundary labels are masked so that a segment's first token is not predicted from the preceding segment.
The per-device packing budgets are 32,768 tokens in Stage~2 and 65,536 tokens in Stage~3, with no padding to a fixed target length.
These physical packing budgets do not change the per-segment context limits of 8K and 64K.

\paragraph{Kernels and numerical precision.}
Our training stack uses VeOmni 0.1.11 \citep{ma2025veomni} and Transformers 5.2.0 with custom LogGate attention modules.
We implement the block scorer and top-$k$ selection in Triton and the sparse attention forward and backward passes in CuTe.
Attention inputs use BF16, while rotary frequencies and position calculations use FP32.
For \dyrope, the implementation uses the intra-block and selected-block rotations described in \S\ref{sec:dyrope} and passes unrotated main queries and keys to that path.
Cached inference stores the corresponding main keys and values and the indexer keys, preserving the same selection and position rules used during training.

\subsection{Evaluation Protocol}
\label{app:exp-evaluation}

\paragraph{General capabilities.}
We use OLMES\footnote{\url{https://github.com/allenai/olmes}} with the task settings in Table~\ref{tab:exp-eval}, applied to the final checkpoint of each stage being evaluated.
Multiple-choice and continuation-ranking tasks use likelihood-based scoring; math, code, and BBH use the generation settings shown in the table.
Math and code report pass@1 estimated from all generated samples, rather than selecting the best sample.
For GSM8K, we remove the double-newline stop string so that an answer beginning with a blank line is not truncated before its content is generated.

\begin{table}[htbp]
\centering
\caption{\textbf{General benchmark evaluation settings.} Samples denotes generations per problem; a dash denotes likelihood-based evaluation. Sampling uses temperature 0.6 and top-$p$ 0.6. All scores are reported on a 0--100 scale.}
\label{tab:exp-eval}
{\tablelargesize
\setlength{\tabcolsep}{5pt}
\renewcommand{\arraystretch}{1.12}
\begin{tabularx}{\linewidth}{l c c l X}
\toprule
Benchmark & Shots & Samples & Evaluation & Reported metric \\
\midrule
GSM8K & 8 & 8 & Sampling & Pass@1 \\
Minerva MATH & 4 & 4 & Sampling & Pass@1 \\
MATH-500 & 4 & 32 & Sampling & Pass@1 \\
MMLU / MMLU-Pro & 5 & --- & Likelihood & Accuracy \\
BBH & 3 & 1 & Sampling & Exact match \\
HellaSwag & 5 & --- & Likelihood & Length-normalized accuracy \\
ARC-C / WinoGrande & 5 & --- & Likelihood & Accuracy \\
DROP & 5 & 1 & Greedy & F1 \\
HumanEval & 3 & 32 & Sampling & Execution pass@1 \\
MBPP & 3 & 32 & Sampling & Execution pass@1 \\
\bottomrule
\end{tabularx}}
\end{table}

Generation limits are 512 tokens for GSM8K, BBH, and MBPP; 1,024 for Minerva MATH, MATH-500, and HumanEval; and 100 for DROP.
Task-specific prompts and other stopping rules follow the stored OLMES configurations.
The Stage-2 Avg.\ in Table~\ref{tab:stage2} is the arithmetic mean over its 15 reported benchmark scores.

\paragraph{Long-context evaluation.}
RULER uses greedy decoding and reports the unweighted mean over its 13 tasks at each context length.
Through 128K, HELMET uses greedy decoding with task-specific generation limits and reports the unweighted mean over its seven categories: recall, retrieval-augmented generation, reranking, in-context learning, citation, long-document question answering, and summarization.
We evaluate the same sampled questions across models at each length, fixing the RAG sampling seed to 42.
Long-context inference uses batch size one and the base-model prompts without a chat template.

\paragraph{HELMET at 256K.}
The standard HELMET configurations cover context lengths only up to 128K, so evaluating 256K extrapolation (Table~\ref{tab:stage3-extrapolation}) requires constructing a 256K extension rather than merely raising the runtime context limit.
Using the OLMo-3 tokenizer, we construct prompts that fill at least 95\% of the 256K input budget after reserving generation tokens: recall contexts are regenerated, demonstration counts and citation retrieval depth are increased, and long-document QA and summarization are restricted to sufficiently long source documents.
However, we exclude RAG and reranking since their released retrieval pools cannot reach this length; across inspected instances, even including all available candidates yields approximately 111K--133K OLMo-3 tokens for RAG and 72K--90K for reranking.
The resulting 256K extension spans five categories across 16 subsets and 1,406 instances (400 recall, 500 in-context learning, 200 citation, 242 long-document QA, and 64 summarization).
All models are evaluated on this same extension, and Table~\ref{tab:stage3-extrapolation} reports their equally weighted category mean.

\paragraph{HELMET scoring.}
We evaluate HELMET through OLMES with targeted fixes to reduce formatting-related evaluation errors.
In the unmodified pipeline, leading whitespace can cause multiple-choice parsing in long-document QA to fail, answer extraction can discard valid content before NarrativeQA judging, and newline-based stopping can terminate RAG generation before an answer is produced.
We therefore tolerate leading whitespace in multiple-choice parsing and judge NarrativeQA on the full generated continuation.
For RAG, we regenerate answers with newline-based stopping disabled while retaining greedy decoding, the model EOS token, and the original 20-token generation budget; we then strip leading whitespace and score only the first remaining line using the original answer-matching metric.
These parsing and stopping rules are applied identically across all evaluated models; our reported HELMET scores therefore follow this modified protocol rather than the unmodified OLMES pipeline and are not directly comparable to unmodified OLMES numbers.
For NarrativeQA and summarization, we use the DeepSeek-V4-Pro judge \citep{xu2026deepseekv4-dsa} with temperature 0.1 and a maximum output of 4,096 tokens, retaining the HELMET grading templates distributed with OLMES: fluency and correctness for NarrativeQA, and fluency, precision, and recall for summarization.

\FloatBarrier
\subsection{Detailed Long-Context Results}
\label{app:long-context-details}
Tables~\ref{tab:stage3-ruler-details}--\ref{tab:stage3-helmet256k-details} give task and category scores for the final Stage-3 checkpoints at step 11{,}921.
Model names follow the main text: Full-RoPE and Sparse-RoPE use YaRN in Stage~3.
All scores are multiplied by 100 and displayed to four decimal places; totals are computed before rounding.

\input{sec/tables/appendix_stage3_ruler_details}
\input{sec/tables/appendix_stage3_helmet_details}
\input{sec/tables/appendix_stage3_helmet256k_details}
\FloatBarrier

\section{Attention Sparsification with an MSA-Style Indexer}
\label{app:msa-conversion}

Before adopting LogGate, we attempted an MSA-style conversion of OLMo-3 to block-sparse attention \citep{lai2026msa}.
The purpose was to obtain a sparse backbone competitive with full attention before introducing \dyrope.
Our trials did not establish the desired parity under the OLMo-3 continued-pretraining setting, so we use LogGate for the positional-encoding comparisons.

\subsection{Reference Indexer and Distillation Objective}
\label{app:msa-reference}

We first describe the reference MSA design, separating it from the implementation-specific choices in our OLMo-3 port.
MSA adds an index-query projection for each KV group and a shared index-key projection.
For hidden state $\vx_i$, index dimension $d_{\mathrm{idx}}$, and routing group $r$, its token score and block score are
\begin{equation}
\begin{aligned}
\vq_i^{\mathrm{idx},r} &= \operatorname{sg}(\vx_i)\mW_q^{\mathrm{idx},r},
&\vk_j^{\mathrm{idx}} &= \operatorname{sg}(\vx_j)\mW_k^{\mathrm{idx}},\\
u_{ij}^{r} &= \frac{\langle\vq_i^{\mathrm{idx},r},\vk_j^{\mathrm{idx}}\rangle}{\sqrt{d_{\mathrm{idx}}}},
&s_{ic}^{r} &= \max_{j\in\gA_{\mathrm{block}}(c),\,j\le i}u_{ij}^{r},
\end{aligned}
\label{eq:msa-indexer}
\end{equation}
where $\operatorname{sg}$ denotes stop-gradient; a block with no causally visible keys has score $-\infty$.
The selector ranks blocks by these max-pooled raw scores, while the main branch performs ordinary softmax attention over the selected, causally valid keys.
The index branch is a pure selector: it does not contribute a separate value aggregation to the layer output.

In the reference design, the query's own block is always retained and occupies one of the total top-block slots; no additional sink block or fixed local window is forced.
This budget convention must be distinguished from the $k$ historical blocks \emph{plus} the self-block used in \S\ref{app:loggate-correction}.
MSA is formulated for GQA, whereas our backbone uses MHA; matching its routing groups to OLMo-3 heads is therefore an implementation choice, not an automatic consequence of adopting its loss.

Let $\mathcal U_i^r$ be the token support used for distillation, and let $\mathcal H_r$ be the set of main-attention heads sharing routing group $r$.
Writing $a_{ij}^{h}$ for the main-branch attention logit, the student and teacher distributions are
\begin{equation}
\begin{aligned}
P_{ij}^{\mathrm{idx},r}
&=\frac{\exp(u_{ij}^{r})}{\sum_{v\in\mathcal U_i^r}\exp(u_{iv}^{r})},\\
P_{ij}^{\mathrm{main},r}
&=\frac{1}{|\mathcal H_r|}\sum_{h\in\mathcal H_r}
\frac{\exp(a_{ij}^{h})}{\sum_{v\in\mathcal U_i^r}\exp(a_{iv}^{h})},
\qquad j\in\mathcal U_i^r.
\end{aligned}
\label{eq:msa-distributions}
\end{equation}
The teacher averages \emph{probabilities}, not logits, and is the model's current main branch rather than a separate frozen dense model.
The KL direction is teacher-to-indexer:
\begin{equation}
\mathcal L_{\mathrm{KL}}
=\frac{1}{|\mathcal Q|R}\sum_{i\in\mathcal Q}\sum_{r=1}^{R}
D_{\mathrm{KL}}\!\left(
\operatorname{sg}(P_{i\cdot}^{\mathrm{main},r})
\,\middle\|\,P_{i\cdot}^{\mathrm{idx},r}\right),
\label{eq:msa-kl}
\end{equation}
where $\mathcal Q$ contains the valid training queries and $R$ is the number of routing groups.
Stop-gradient on the teacher and on the indexer's hidden-state inputs confines the KL gradient to the index projections.
It does \emph{not} imply that the whole backbone is frozen: backbone updates from the language-modeling loss are a separate choice.

The reference training has two stages.
During \emph{indexer warmup}, the main branch remains dense and $\mathcal U_i^r=\{0,\ldots,i\}$, avoiding routing by an untrained indexer.
During \emph{sparse training}, the main branch switches to the selected blocks and $\mathcal U_i^r$ is their causally valid token union.
Both distributions in \eqref{eq:msa-distributions} are then renormalised on that selected support; the teacher is not an unnormalised slice of dense attention.
The reference total objective combines the language-modeling loss with layerwise distillation,
\begin{equation}
\mathcal L=\mathcal L_{\mathrm{LM}}+
\lambda\sum_{\ell\in\mathcal L_{\mathrm{sparse}}}\mathcal L_{\mathrm{KL}}^{(\ell)}.
\label{eq:msa-objective}
\end{equation}
The discrete top-$k$ indices are not differentiated, so the KL path, rather than the language-modeling loss through routing, trains the indexer.

\subsection{MSA-KL Conversion Experiments on OLMo-3}
\label{app:msa-kl}

We refer to the conversion that trains the indexer of \eqref{eq:msa-indexer} with the objective in \eqref{eq:msa-objective} as MSA-KL, and report its runs at both OLMo-3 training stages: 8K mid-training (\S\ref{app:msa-kl-8k}) and 64K long-context extension (\S\ref{app:msa-kl-64k}).
Unless specified below, the data, tokenizer, optimizer, and evaluation settings follow the Full Attn.\ configuration of the corresponding stage in Appendix~\ref{app:experimental-details}. We detail the MSA-KL implementation, indexer warmup, and route-specific initialization and training budgets below.

\paragraph{Implementation.}
We replace the eight global-attention layers with MSA layers and retain the 24 sliding-window layers with a window size of 4,096.
Each MSA layer has four routing groups of eight heads: each group has its own index query, all groups share one index key, $d_{\mathrm{idx}}=128$, and index queries and keys are rotated by the main branch's RoPE.
For each query and group, the selector keeps 16 blocks of $b=128$ tokens \emph{including} the query's own block, i.e.\ at most 2,048 keys, as in MSA.
With the stop-gradients of \S\ref{app:msa-reference}, the KL updates only the index projections, while the language-modeling loss trains the backbone in both the warmup and the sparse stage.
We set $\lambda=1$ on the \emph{mean} KL over the eight MSA layers, i.e.\ $\lambda=1/8$ in the layer-sum form of \eqref{eq:msa-objective}.

\subsubsection{8K Mid-Training}
\label{app:msa-kl-8k}

\paragraph{Setup.}
MSA-KL is converted from the same OLMo-3-7B pre-training checkpoint as Full Attn.\ in Table~\ref{tab:stage2} and trained on the same 50B tokens of 8K mid-training data, in 24,000 steps of 256 sequences.
Following the warmup share of MSA's continued pretraining (40B of 400B tokens), the first 2,400 steps (5.0B tokens) are indexer warmup and the remaining 21,600 steps (45.3B tokens) sparse training; the 50B budget thus \emph{includes} the warmup.
MSA does not report the learning rates of its warmup and sparse stages.
For a controlled comparison in the OLMo-3 setting, we therefore keep every other setting of the Full Attn.\ run, including its RoPE ($\theta=500{,}000$, unscaled) and its learning rate: one schedule spans both stages, warming up for 199 steps to $2.07\times10^{-4}$ and decaying linearly to zero.

\input{sec/tables/appendix_msa_kl_8k}

\paragraph{Results.}
MSA-KL trails Full Attn.\ by 1.0 point on the 15-benchmark average (Table~\ref{tab:msa-kl-8k}), but the deficit is far from uniform.
On the 12 math, general and code benchmarks the two means differ by only 0.2 points: MSA-KL matches or exceeds Full Attn.\ on all three math tasks, stays within 1.3 points on the general ones, and loses most on HumanEval ($-2.5$).
The rest of the gap lies in long context: RULER is on par at 4K ($+0.5$) but drops by 4.8 points at the 8K training length, and HELMET by 8.0 points.
At 4K the sliding-window layers alone span the whole context; beyond 4K, long-range retrieval must pass through the global layers, where each query sees at most 2,048 selected keys.
The pattern is therefore consistent with block selection, rather than the general ability of the converted backbone, limiting MSA-KL.

\subsubsection{64K Long-Context Extension}
\label{app:msa-kl-64k}

\paragraph{Setup.}
For long context, MSA states only that its extension continues the sparse continued-pretraining checkpoint for about 140B tokens; the training length, learning rate and positional scaling are not given, nor is whether the indexer warmup is repeated.
To realise a 64K MSA-KL extension on the OLMo-3 architecture and its 64K continued-pretraining (CPT) data, we therefore compare four initialisation routes (Table~\ref{tab:msa-kl-64k}):
(A)~MSA-KL 8K CPT (24K steps, about 50B tokens; \S\ref{app:msa-kl-8k}) $\rightarrow$ sparse KL training on the 64K CPT data, without repeating the indexer warmup;
(B)~Full Attn.\ 64K CPT after 4K steps $\rightarrow$ indexer warmup and sparse KL training on the remaining 64K CPT data, about 3.3B and 30B tokens respectively;
(C)~the complete Full Attn.\ 64K CPT checkpoint $\rightarrow$ indexer warmup and sparse KL training on a replay of the 64K CPT data;
(D)~Full Attn.\ 8K CPT (24K steps, about 50B tokens; Table~\ref{tab:stage2}) $\rightarrow$ indexer warmup and sparse KL training on the 64K CPT data, with a 512-token local window alongside the 16 selected blocks.
The full-attention 64K CPT that B and C start from is an earlier run with the same 64K recipe but a different 8K predecessor, whose learning rate was decayed to $10^{-4}$ rather than zero; it is therefore not the Full Attn.\ run of Table~\ref{tab:stage3}, and reaches 68.30 instead of 71.95 on RULER at 64K.
For a fair comparison in the OLMo-3 setting, all routes otherwise follow Full Attn.: they use the same 64K data and batch (64 sequences of 64K tokens) and the same positional encoding, YaRN with factor 8 over the 8K Stage-2 length in the eight global layers, applied to both main and index queries and keys.
They also share its learning-rate schedule: 200 warmup steps to $2.07\times10^{-4}$, then linear decay to zero at the end of each route.

\input{sec/tables/appendix_msa_kl_64k_ruler}

\paragraph{RULER.}
All four routes follow the same pattern (Table~\ref{tab:msa-kl-64k}): they are on par with Full Attn.\ up to 8K (between $-2.72$ and $+1.34$ points), after which the gap widens with length, to 8.8--11.3 points at 16K, 15.7--22.3 at 32K and 23.9--28.3 at 64K.
The choice of route matters far less than this gap: averaged over 16K--64K, the routes lie within 3.2 points of each other (58.57--61.73), but at least 17.1 points below Full Attn.
Repeating the warmup at 64K (B--D) helps mainly at 32K, where it gains 4.3--6.7 points over route A, which reuses the 8K indexer; replaying the data (C) or widening the local window (D) does not raise the 16K--64K average above that of B, which also scores best at 64K among the routes (48.06).

\input{sec/tables/appendix_msa_kl_64k_helmet}

\paragraph{HELMET.}
We evaluate the best route, B, on HELMET (Table~\ref{tab:msa-kl-helmet}).
It is within 1.97 points of Full Attn.\ at 8K, but the gap grows to 6.92, 10.51 and 13.70 points at 16K, 32K and 64K; from 8K to 64K, Full Attn.\ retains 90\% of its HELMET score, MSA-KL only 64\%.
The realistic tasks of HELMET thus confirm the RULER trend: the deficit of MSA-KL grows steadily with context length.

\paragraph{Summary.}
In the evaluated OLMo-3 conversions, MSA-KL retains short-context capability, while its RULER score falls 4.8 points below Full Attn.\ at 8K after mid-training and at least 23.9 points below at 64K after extension. LogGate provides a competitive sparse backbone under the main-experiment recipe (Tables~\ref{tab:stage2} and~\ref{tab:stage3}), motivating its use for the positional-encoding comparisons in \S\ref{sec:loggate}.
\section{\acronym Overhead: Measurement and Analysis}
\label{app:efficiency-measurement}
\paragraph{Measurement and calibration.}
The paired PE-off/on runs share sparse selections and gates, with both Q/K pre-rotations outside timing.
After compilation and warmup, CUDA Graph replay and same-stream GPU events avoid per-operator Python dispatch gaps; core-only graphs exclude indexer and union construction.
Each workload uses ten alternating AB/BA rounds, taking each variant's median before computing the increment; 95\% bootstrap intervals resample five independent seeds or ten documents.
We retain the production CuTe PE-dependent pipeline settings, so the difference includes the execution effects of enabling the feature.
Graph contents, output agreement, and GPU exclusivity are checked.
For synthetic workloads, each 128-row tile packs $128/G$ tokens with $G$ Q heads per token.
All four $G$ values share per-query selections; three separate calibration seeds fit the real mean unions over 16/32/64/128-token windows at each length.
Formal-seed mean-union errors are at most 5.90\% over the full sequence and 10.85\% over its second half; this matches selected statistics, not every access pattern or a real GQA checkpoint.

\begin{figure}[!htbp]
\centering
\includegraphics[width=0.78\textwidth]{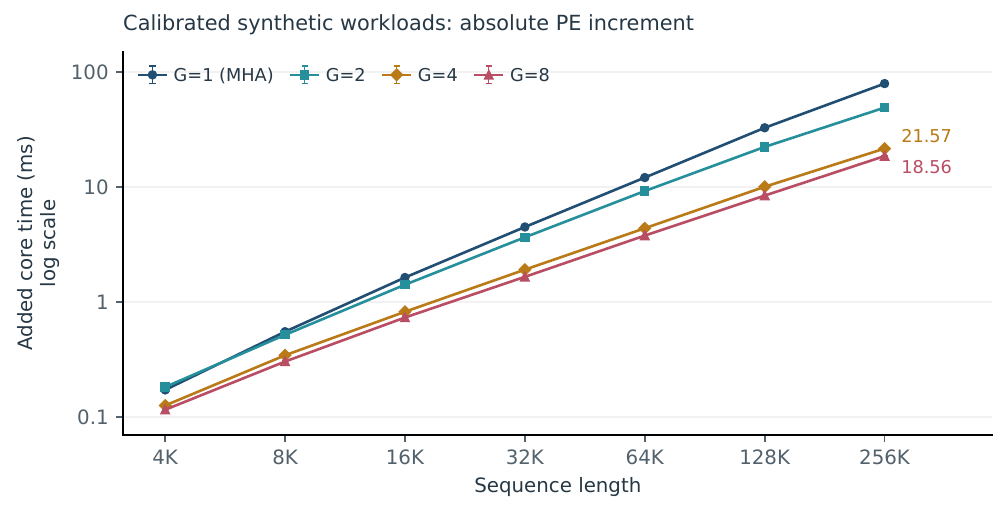}
\caption{Absolute forward PE increments for the calibrated synthetic workloads in Figure~\ref{fig:dyrope-efficiency}(b).
Each point is the mean additional time per core layer over five independent seeds; thin error bars are 95\% bootstrap intervals.
The vertical axis is logarithmic, with tick labels in milliseconds.}
\label{fig:dyrope-absolute-ms}
\end{figure}

\paragraph{Absolute increments under GQA.}
Across 8K--256K, the absolute additional time per core layer decreases with GQA group size (Figure~\ref{fig:dyrope-absolute-ms}).
At 256K, $G=8$ adds 18.56 ms versus 21.57 ms for $G=4$ and 26.54 ms for MHA.
Because the PE-off baseline shrinks faster with $G$ than the increment does, a smaller absolute cost can coincide with a higher percentage overhead.

\begin{figure}[!htbp]
\centering
\includegraphics[width=\textwidth]{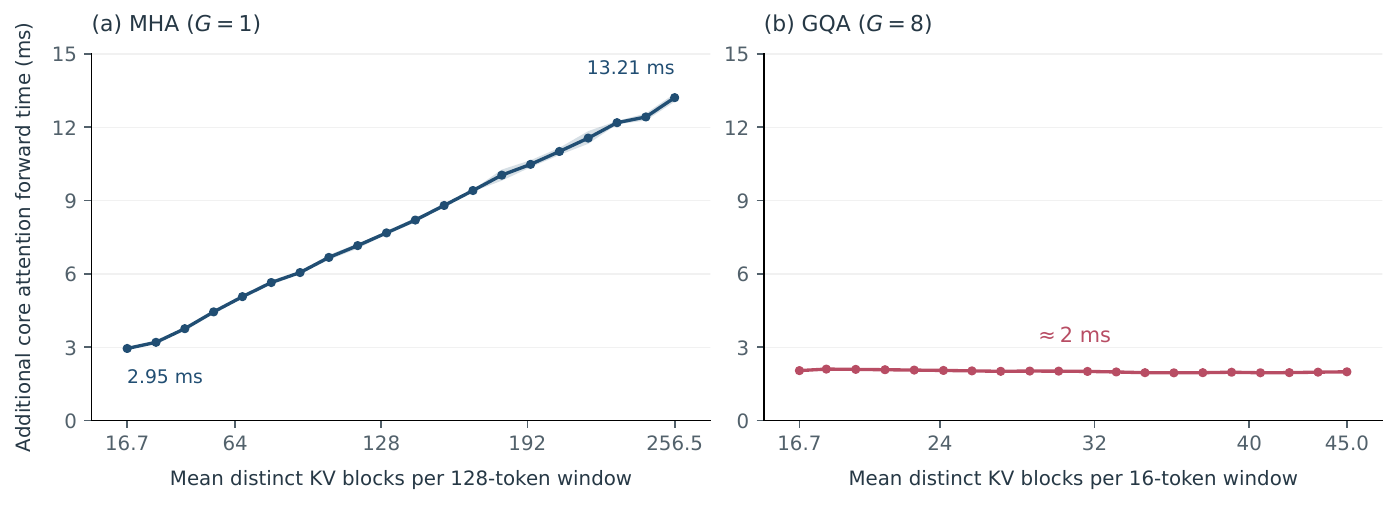}
\caption{Selection-overlap sweep at 64K with a fixed per-query budget and smoothly varying selections.
The horizontal axes count distinct selected KV blocks averaged over 128-token windows for MHA and 16-token windows for GQA.
Additional time is $T_{\mathrm{on}}-T_{\mathrm{off}}$ per core attention layer.
Shading shows 95\% bootstrap intervals over five seeds.}
\label{fig:dyrope-union-sweep}
\end{figure}

\paragraph{Sensitivity to selection overlap.}
Figure~\ref{fig:dyrope-union-sweep} varies selection overlap at a fixed 64K length and a fixed per-query budget of up to 16 historical blocks plus the current block.
We enlarge a candidate pool and slide each query's selections smoothly through it.
Each panel measures the mean number of distinct selected KV blocks per kernel window: 128 tokens for MHA ($G=1$) and 16 tokens for GQA ($G=8$).
Across twenty settings, these means span 16.73--256.50 and 16.73--45.05, respectively.
The additional forward time rises from 2.95 to 13.21 ms for MHA and remains approximately 2 ms for GQA under this smooth selection pattern.
Larger unions require core attention to process more blocks, increasing its computation and baseline runtime; the baseline grows proportionally faster than the \acronym increment.
Consequently, the relative overhead decreases as the union grows: from 13.54\% (mean union 16.73) to 4.80\% (256.50) for MHA, and from 9.85\% (16.73) to 3.96\% (45.05) for GQA.

%% file: sec/tables/appendix_stage3_ruler_details.tex
\begingroup
\tablelargesize
\setlength{\tabcolsep}{7pt}
\renewcommand{\arraystretch}{1.10}
\setlength{\LTcapwidth}{\linewidth}
\begin{longtable}{@{}l*{4}{r}>{\columncolor{oursbg}}r@{}}
\caption{\textbf{Stage-3 RULER scores by task.} 64K is the training length; 128K, 256K, and 512K are extrapolation settings. Each task has 100 examples per model and length; Total averages the 13 tasks equally. Lengths denote target input budgets. Best in each row is bold; \dyrope is shaded.}\label{tab:stage3-ruler-details} \\
\toprule
Task & Full-RoPE & Sparse-RoPE & Sparse-HoPE & Sparse-NoPE & Sparse-\dyrope \\
\midrule
\endfirsthead
\caption[]{\textbf{Stage-3 RULER scores by task (continued).}} \\
\toprule
Task & Full-RoPE & Sparse-RoPE & Sparse-HoPE & Sparse-NoPE & Sparse-\dyrope \\
\midrule
\endhead
\midrule
\multicolumn{6}{r}{\emph{Continued on the next page.}} \\
\endfoot
\bottomrule
\endlastfoot
\rowcolor{groupbg}\multicolumn{6}{l}{\textbf{64K} (training length)} \\*
NIAH-S1 & \tbest{100.0000} & \tbest{100.0000} & \tbest{100.0000} & \tbest{100.0000} & \tbest{100.0000} \\*
NIAH-S2 & \tbest{100.0000} & \tbest{100.0000} & \tbest{100.0000} & \tbest{100.0000} & \tbest{100.0000} \\*
NIAH-S3 & \tbest{100.0000} & \tbest{100.0000} & 99.0000 & \tbest{100.0000} & \tbest{100.0000} \\*
NIAH-MK1 & 83.0000 & 75.0000 & 86.0000 & \tbest{91.0000} & 87.0000 \\*
NIAH-MK2 & 88.0000 & \tbest{98.0000} & 92.0000 & 97.0000 & 97.0000 \\*
NIAH-MK3 & 88.0000 & 85.0000 & 65.0000 & 91.0000 & \tbest{98.0000} \\*
NIAH-MV & 79.2500 & 75.5000 & 91.7500 & 86.2500 & \tbest{92.7500} \\*
NIAH-MQ & 85.2500 & 80.5000 & 90.5000 & \tbest{95.5000} & 91.2500 \\*
VT & 34.0000 & 47.2000 & 6.4000 & 40.0000 & \tbest{68.8000} \\*
CWE & 11.9000 & 16.4000 & 16.2000 & \tbest{17.8000} & 15.7000 \\*
FWE & 68.0000 & 79.6667 & 83.0000 & \tbest{90.6667} & 89.6667 \\*
QA-1 & 58.0000 & 61.0000 & 55.0000 & \tbest{68.0000} & 63.0000 \\*
QA-2 & 40.0000 & 45.0000 & 43.0000 & \tbest{48.0000} & 40.0000 \\*
\cmidrule(lr){1-6}
\textbf{Total} & 71.9538 & 74.0974 & 71.3731 & 78.8628 & \tbest{80.2436} \\
\addlinespace[6pt]
\rowcolor{groupbg}\multicolumn{6}{l}{\textbf{128K} (extrapolation)} \\*
NIAH-S1 & 46.0000 & 86.0000 & \tbest{100.0000} & \tbest{100.0000} & \tbest{100.0000} \\*
NIAH-S2 & 23.0000 & 54.0000 & \tbest{100.0000} & 98.0000 & 97.0000 \\*
NIAH-S3 & 13.0000 & 67.0000 & 99.0000 & 98.0000 & \tbest{100.0000} \\*
NIAH-MK1 & 20.0000 & 32.0000 & 74.0000 & \tbest{90.0000} & 78.0000 \\*
NIAH-MK2 & 2.0000 & 26.0000 & 91.0000 & 94.0000 & \tbest{96.0000} \\*
NIAH-MK3 & 0.0000 & 18.0000 & 48.0000 & \tbest{90.0000} & 89.0000 \\*
NIAH-MV & 14.7500 & 37.0000 & 88.7500 & 86.2500 & \tbest{94.0000} \\*
NIAH-MQ & 11.5000 & 33.7500 & 94.0000 & 95.7500 & \tbest{96.5000} \\*
VT & 3.8000 & 48.6000 & 3.8000 & 27.8000 & \tbest{60.2000} \\*
CWE & 1.0000 & \tbest{4.3000} & 2.3000 & 3.5000 & 1.5000 \\*
FWE & 3.6667 & 77.3333 & \tbest{95.6667} & 89.0000 & \tbest{95.6667} \\*
QA-1 & 11.0000 & 36.0000 & 47.0000 & \tbest{67.0000} & 56.0000 \\*
QA-2 & 20.0000 & 29.0000 & 41.0000 & 38.0000 & \tbest{44.0000} \\*
\cmidrule(lr){1-6}
\textbf{Total} & 13.0551 & 42.2295 & 68.0397 & 75.1769 & \tbest{77.5282} \\
\pagebreak[4]
\rowcolor{groupbg}\multicolumn{6}{l}{\textbf{256K} (extrapolation)} \\*
NIAH-S1 & 17.0000 & 49.0000 & \tbest{100.0000} & \tbest{100.0000} & \tbest{100.0000} \\*
NIAH-S2 & 7.0000 & 26.0000 & \tbest{100.0000} & \tbest{100.0000} & 99.0000 \\*
NIAH-S3 & 0.0000 & 24.0000 & 99.0000 & \tbest{100.0000} & \tbest{100.0000} \\*
NIAH-MK1 & 6.0000 & 25.0000 & 72.0000 & \tbest{90.0000} & 87.0000 \\*
NIAH-MK2 & 0.0000 & 14.0000 & 55.0000 & 68.0000 & \tbest{83.0000} \\*
NIAH-MK3 & 0.0000 & 12.0000 & 23.0000 & \tbest{75.0000} & 72.0000 \\*
NIAH-MV & 2.5000 & 19.2500 & 75.0000 & 76.2500 & \tbest{88.2500} \\*
NIAH-MQ & 2.0000 & 14.5000 & 84.2500 & 87.0000 & \tbest{89.2500} \\*
VT & 0.0000 & 9.2000 & 1.0000 & 42.0000 & \tbest{68.2000} \\*
CWE & 0.1000 & 0.3000 & 0.2000 & \tbest{1.1000} & 0.1000 \\*
FWE & 18.0000 & 67.6667 & 88.6667 & 86.3333 & \tbest{92.6667} \\*
QA-1 & 7.0000 & 17.0000 & 42.0000 & \tbest{69.0000} & 56.0000 \\*
QA-2 & 13.0000 & 25.0000 & 37.0000 & \tbest{47.0000} & 42.0000 \\*
\cmidrule(lr){1-6}
\textbf{Total} & 5.5846 & 23.3013 & 59.7782 & 72.4372 & \tbest{75.1897} \\
\addlinespace[6pt]
\rowcolor{groupbg}\multicolumn{6}{l}{\textbf{512K} (extrapolation)} \\*
NIAH-S1 & 3.0000 & 25.0000 & \tbest{100.0000} & \tbest{100.0000} & \tbest{100.0000} \\*
NIAH-S2 & 0.0000 & 9.0000 & \tbest{100.0000} & 97.0000 & 84.0000 \\*
NIAH-S3 & 1.0000 & 14.0000 & \tbest{99.0000} & \tbest{99.0000} & \tbest{99.0000} \\*
NIAH-MK1 & 2.0000 & 13.0000 & 72.0000 & \tbest{91.0000} & 79.0000 \\*
NIAH-MK2 & 0.0000 & 6.0000 & 46.0000 & 54.0000 & \tbest{78.0000} \\*
NIAH-MK3 & 0.0000 & 5.0000 & 12.0000 & \tbest{71.0000} & 62.0000 \\*
NIAH-MV & 0.2500 & 8.5000 & 79.5000 & 81.0000 & \tbest{89.0000} \\*
NIAH-MQ & 1.5000 & 8.5000 & 76.0000 & 80.0000 & \tbest{83.5000} \\*
VT & 0.0000 & 5.8000 & 1.0000 & 44.4000 & \tbest{72.0000} \\*
CWE & 0.1000 & 0.3000 & 0.2000 & \tbest{1.1000} & 0.1000 \\*
FWE & 9.6667 & 60.6667 & 85.3333 & 85.6667 & \tbest{94.6667} \\*
QA-1 & 7.0000 & 17.0000 & 42.0000 & \tbest{69.0000} & 56.0000 \\*
QA-2 & 12.0000 & 25.0000 & 27.0000 & \tbest{45.0000} & 41.0000 \\*
\cmidrule(lr){1-6}
\textbf{Total} & 2.8090 & 15.2128 & 56.9256 & 70.6282 & \tbest{72.1744} \\
\end{longtable}
\endgroup

%% file: sec/tables/appendix_stage3_helmet_details.tex
\begin{table}[!htbp]
\centering
\caption{\textbf{Stage-3 HELMET scores by category.} Each length contains 21 tasks and 2,100 examples per model. Tasks are averaged equally within each category; Total averages the seven categories equally. 128K exceeds the 64K training length. Best per category and Total at each length is bold; \dyrope is shaded.}
\label{tab:stage3-helmet-details}
\begingroup
\tablesize
\setlength{\tabcolsep}{3pt}
\renewcommand{\arraystretch}{1.16}
\begin{tabularx}{\linewidth}{@{}l*{8}{>{\raggedleft\arraybackslash}X}@{}}
\toprule
Model & Recall & RAG & Rerank & ICL & Cite & LongQA & Summ & Total \\
\midrule
\rowcolor{groupbg}\multicolumn{9}{l}{\textbf{8K}} \\
Full-RoPE & 98.6250 & 54.5000 & 49.0830 & \tbest{72.2000} & \tbest{21.6617} & 37.9110 & 1.8516 & 47.9760 \\
Sparse-RoPE & 98.6250 & \tbest{58.2500} & 44.9475 & 69.0000 & 20.2917 & \tbest{44.6977} & 4.1135 & \tbest{48.5608} \\
Sparse-HoPE & \tbest{98.8125} & 52.0000 & 38.2636 & 71.0000 & 20.3395 & 40.7630 & 1.5226 & 46.1002 \\
Sparse-NoPE & 97.6250 & 58.0000 & \tbest{52.8661} & 67.2000 & 20.1804 & 35.7782 & 3.0253 & 47.8107 \\
\rowcolor{oursbg}Sparse-\dyrope & 98.6250 & 56.0000 & 48.8513 & 68.2000 & 18.7104 & 40.6871 & \tbest{5.5777} & 48.0931 \\
\addlinespace[5pt]
\rowcolor{groupbg}\multicolumn{9}{l}{\textbf{16K}} \\
Full-RoPE & 96.8125 & 53.5000 & 32.6715 & \tbest{80.4000} & 14.7851 & \tbest{49.6014} & 1.9338 & 47.1006 \\
Sparse-RoPE & \tbest{98.9375} & 57.7500 & 26.0354 & 78.2000 & 16.8995 & 49.5305 & 3.5849 & 47.2768 \\
Sparse-HoPE & 93.6875 & 54.2500 & 28.9196 & 76.0000 & 12.4934 & 47.4824 & 1.4897 & 44.9032 \\
Sparse-NoPE & 96.9375 & \tbest{59.0000} & \tbest{38.5204} & 77.4000 & 18.2783 & 43.7904 & 2.5285 & \tbest{48.0650} \\
\rowcolor{oursbg}Sparse-\dyrope & 97.5625 & 54.7500 & 30.9462 & 76.2000 & \tbest{18.5571} & 48.6334 & \tbest{5.2525} & 47.4145 \\
\addlinespace[5pt]
\rowcolor{groupbg}\multicolumn{9}{l}{\textbf{32K}} \\
Full-RoPE & 96.7500 & 49.2500 & 20.9059 & \tbest{82.0000} & 10.8274 & 52.0262 & 1.5204 & 44.7543 \\
Sparse-RoPE & \tbest{97.2500} & 56.7500 & 20.0153 & \tbest{82.0000} & 9.0476 & 59.0974 & 3.2673 & 46.7754 \\
Sparse-HoPE & 89.5625 & 53.7500 & 18.8995 & 81.2000 & 11.7157 & \tbest{59.6109} & 1.2431 & 45.1402 \\
Sparse-NoPE & 95.0000 & \tbest{57.0000} & \tbest{28.7761} & 80.6000 & 14.4979 & 58.7302 & 2.3295 & \tbest{48.1334} \\
\rowcolor{oursbg}Sparse-\dyrope & 96.9375 & 54.5000 & 15.7840 & 81.4000 & \tbest{14.8400} & 57.1169 & \tbest{3.8791} & 46.3511 \\
\addlinespace[5pt]
\rowcolor{groupbg}\multicolumn{9}{l}{\textbf{64K} (training length)} \\
Full-RoPE & 87.7500 & 48.5000 & 7.7766 & 85.6000 & 5.9010 & 65.8580 & 1.8459 & 43.3188 \\
Sparse-RoPE & 85.4375 & 51.7500 & 6.9278 & 84.4000 & 7.3148 & 62.2987 & 2.6587 & 42.9696 \\
Sparse-HoPE & 81.5625 & 54.7500 & 9.6767 & 84.0000 & 6.7933 & 68.4901 & 1.9068 & 43.8828 \\
Sparse-NoPE & 92.5000 & \tbest{55.2500} & \tbest{24.7336} & 85.4000 & 13.1112 & \tbest{70.7250} & 2.6487 & \tbest{49.1955} \\
\rowcolor{oursbg}Sparse-\dyrope & \tbest{95.4375} & 50.2500 & 11.2886 & \tbest{86.0000} & \tbest{15.7219} & 69.7671 & \tbest{3.1146} & 47.3685 \\
\addlinespace[5pt]
\rowcolor{groupbg}\multicolumn{9}{l}{\textbf{128K} (extrapolation)} \\
Full-RoPE & 4.3125 & 20.7500 & 0.0000 & 78.2000 & 2.3333 & 46.0760 & 12.5040 & 23.4537 \\
Sparse-RoPE & 23.9375 & 35.5000 & 0.0000 & 83.2000 & 2.2083 & 52.2116 & \tbest{15.8028} & 30.4086 \\
Sparse-HoPE & 74.0625 & 50.7500 & 6.7095 & 85.8000 & 7.4030 & 70.1331 & 1.7996 & 42.3797 \\
Sparse-NoPE & 90.5625 & \tbest{54.0000} & \tbest{16.9551} & 86.4000 & 5.0567 & 69.7246 & 2.4762 & 46.4536 \\
\rowcolor{oursbg}Sparse-\dyrope & \tbest{91.8750} & 52.5000 & 8.0644 & \tbest{88.8000} & \tbest{9.8078} & \tbest{72.4471} & 2.7645 & \tbest{46.6084} \\
\bottomrule
\end{tabularx}
\endgroup
\end{table}

%% file: sec/tables/appendix_stage3_helmet256k_details.tex
\begin{table}[!htbp]
\centering
\caption{\textbf{Stage-3 HELMET-256K subset scores by category.} This extrapolation setting contains 16 tasks and 1,406 examples across five categories. Total gives these categories equal weight, excluding RAG and reranking. Its category coverage and document selection differ from the seven-category evaluation through 128K. Best per column is bold; \dyrope is shaded.}
\label{tab:stage3-helmet256k-details}
\begingroup
\tablelargesize
\setlength{\tabcolsep}{3pt}
\renewcommand{\arraystretch}{1.16}
\begin{tabularx}{\linewidth}{@{}l*{6}{>{\raggedleft\arraybackslash}X}@{}}
\toprule
Model & Recall & ICL & Cite & LongQA & Summ & Total \\
\midrule
Full-RoPE & 1.0625 & 70.4000 & 0.6250 & 14.4014 & 0.0000 & 17.2978 \\
Sparse-RoPE & 13.1875 & 81.2000 & 0.9583 & 41.1349 & 3.8871 & 28.0736 \\
Sparse-HoPE & 51.8125 & 82.4000 & 4.5139 & 48.5950 & \tbest{10.2029} & 39.5048 \\
Sparse-NoPE & 74.8125 & 86.4000 & 2.5833 & 64.1637 & 1.9921 & 45.9903 \\
\rowcolor{oursbg}Sparse-\dyrope & \tbest{79.8750} & \tbest{88.6000} & \tbest{8.5735} & \tbest{75.8053} & 2.8082 & \tbest{51.1324} \\
\bottomrule
\end{tabularx}
\endgroup
\end{table}

%% file: sec/tables/appendix_msa_kl_8k.tex
\begin{table}[t]
\centering
\caption{
  \textbf{MSA-KL at 8K mid-training (Stage~2).}
  Both models start from the OLMo-3-7B pre-training checkpoint of Table~\ref{tab:stage2} and are trained on the same 50B tokens of 8K mid-training data, which for MSA-KL include the 5.0B-token indexer warmup.
  Benchmarks and scoring follow Table~\ref{tab:stage2}; Avg.\ is the mean over the 15 benchmarks, and $\Delta$ is MSA-KL minus Full Attn.
}
\label{tab:msa-kl-8k}
\vspace{4pt}
{\tablesize
\setlength{\tabcolsep}{2.4pt}%
\renewcommand{\arraystretch}{1.18}%
\def\hrotangle{45}%
\def\hrot#1{\makebox[0pt][l]{\rotatebox{\hrotangle}{#1}}}%
\begin{NiceTabular}{l *{16}{c}}
\toprule
 & \multicolumn{3}{c}{\textbf{Math}} & \multicolumn{7}{c}{\textbf{General}} & \multicolumn{2}{c}{\textbf{Code}} & \multicolumn{3}{c}{\textbf{Long context}} & \\
\cmidrule(lr){2-4} \cmidrule(lr){5-11} \cmidrule(lr){12-13} \cmidrule(lr){14-16}
\textbf{Model} & \hrot{GSM8K} & \hrot{Minerva MATH} & \hrot{MATH-500} & \hrot{MMLU} & \hrot{MMLU-Pro} & \hrot{BBH} & \hrot{HellaSwag} & \hrot{ARC-C} & \hrot{WinoGrande} & \hrot{DROP} & \hrot{HumanEval} & \hrot{MBPP} & \hrot{RULER 4K} & \hrot{RULER 8K} & \hrot{HELMET 8K} & \hrot{Avg.} \\
\midrule
Full Attn. & 73.35 & 35.99 & 36.77 & 65.48 & 34.80 & 64.80 & 77.57 & 84.32 & 85.84 & 60.40 & 45.33 & 49.01 & 92.65 & 80.35 & 46.47 & 62.21 \\
MSA-KL & 73.4\phantom{0} & 36.24 & 38.98 & 65.88 & 34.91 & 64.01 & 77.41 & 84.09 & 85.99 & 59.07 & 42.87 & 48.46 & 93.16 & 75.59 & 38.48 & 61.24 \\
\arrayrulecolor{black!30}\midrule\arrayrulecolor{black}
$\Delta$ & $+0.1$ & $+0.3$ & $+2.2$ & $+0.4$ & $+0.1$ & $-0.8$ & $-0.2$ & $-0.2$ & $+0.2$ & $-1.3$ & $-2.5$ & $-0.6$ & $+0.5$ & $-4.8$ & $-8.0$ & $-1.0$ \\
\bottomrule
\end{NiceTabular}
}
\end{table}

%% file: sec/tables/appendix_msa_kl_64k_ruler.tex
\begin{table}[t]
\centering
\caption{
  \textbf{MSA-KL routes to 64K (Stage~3) on RULER.}
  Full Attn.\ is the RoPE + YaRN row of Table~\ref{tab:stage3} (dense, 50B tokens of 64K CPT data); all models use RoPE + YaRN.
  Initialisation is the checkpoint a route starts from (8K or 64K CPT stage); Data states whether the route makes one full pass over the 64K CPT data, uses the part its initialisation has not consumed, or replays it.
  Tokens are nominal budgets for indexer warmup (dense attention) and sparse KL training; Avg.\ is the mean over 16K--64K.
  $^\dagger$An earlier full-attention 64K CPT run with the same recipe, continued from an 8K checkpoint whose learning rate was decayed to $10^{-4}$ instead of zero; not the Full Attn.\ run above.
  $^\ddagger$With a 512-token local window alongside the 16 selected blocks.
  Best per column in \tbest{bold}, second \tsecond{underlined}.
}
\label{tab:msa-kl-64k}
\vspace{4pt}
{\tablesize
\setlength{\tabcolsep}{2.1pt}%
\renewcommand{\arraystretch}{1.18}%
\begin{NiceTabular}{l l l l r r *{5}{c} w{c}{1pt} c}
\toprule
 & & & & \multicolumn{2}{c}{\textbf{Tokens}} & \multicolumn{5}{c}{\textbf{RULER}} & & \\
\cmidrule(lr){5-6} \cmidrule(lr){7-11}
\textbf{Attention} & \textbf{Route} & \textbf{Initialisation} & \textbf{Data} & Warmup & Sparse & 4K & 8K & 16K & 32K & 64K & & Avg. \\
\midrule
Full Attn. & \tna & Full Attn.\ 8K & full & \tna & \tna & 93.87 & \tbest{89.86} & \tbest{84.53} & \tbest{80.04} & \tbest{71.95} & & \tbest{78.84} \\
\arrayrulecolor{black!30}\midrule\arrayrulecolor{black}
\Block[l]{4-1}{MSA-KL}
  & A & MSA-KL 8K & full & 0 & 50.0B & 94.01 & 87.14 & 73.28 & 57.70 & 44.72 & & 58.57 \\
  & B & Full Attn.$^\dagger$ 64K, 4K steps & remaining & 3.3B & 29.9B & \tbest{95.21} & 89.29 & 73.68 & 63.44 & \tsecond{48.06} & & \tsecond{61.73} \\
  & C & Full Attn.$^\dagger$ 64K, complete & replay & 5.0B & 45.0B & 93.12 & \tsecond{89.68} & 75.68 & \tsecond{64.37} & 43.65 & & 61.23 \\
  & D$^\ddagger$ & Full Attn.\ 8K & full & 5.0B & 45.0B & \tsecond{94.56} & 87.36 & \tsecond{75.72} & 61.95 & 45.53 & & 61.07 \\
\bottomrule
\end{NiceTabular}
}
\end{table}

%% file: sec/tables/appendix_msa_kl_64k_helmet.tex
\begin{table}[t]
\centering
\caption{
  \textbf{HELMET of the best MSA-KL route at 64K (Stage~3).}
  Route B of Table~\ref{tab:msa-kl-64k} against the Full Attn.\ row of Table~\ref{tab:stage3}, under the same HELMET protocol.
  Avg.\ is the mean over the four lengths, and $\Delta$ is MSA-KL (B) minus Full Attn.
}
\label{tab:msa-kl-helmet}
\vspace{4pt}
{\tablesize
\setlength{\tabcolsep}{4pt}%
\renewcommand{\arraystretch}{1.18}%
\begin{NiceTabular}{l *{4}{w{c}{1.0cm}} w{c}{1pt} w{c}{1.0cm}}
\toprule
 & \multicolumn{4}{c}{\textbf{HELMET}} & & \\
\cmidrule(lr){2-5}
\textbf{Model} & 8K & 16K & 32K & 64K & & Avg. \\
\midrule
Full Attn. & 47.98 & 47.10 & 44.75 & 43.32 & & 45.79 \\
MSA-KL (B) & 46.01 & 40.18 & 34.24 & 29.62 & & 37.51 \\
\arrayrulecolor{black!30}\midrule\arrayrulecolor{black}
$\Delta$ & $-1.97$ & $-6.92$ & $-10.51$ & $-13.70$ & & $-8.28$ \\
\bottomrule
\end{NiceTabular}
}
\end{table}

%% file: main.bbl
\begin{thebibliography}{58}
\providecommand{\natexlab}[1]{#1}
\providecommand{\url}[1]{\texttt{#1}}
\expandafter\ifx\csname urlstyle\endcsname\relax
  \providecommand{\doi}[1]{doi: #1}\else
  \providecommand{\doi}{doi: \begingroup \urlstyle{rm}\Url}\fi

\bibitem[Ainslie et~al.(2023)Ainslie, Lee-Thorp, de~Jong, Zemlyanskiy, Lebron, and Sanghai]{ainslie2023gqa}
Joshua Ainslie, James Lee-Thorp, Michiel de~Jong, Yury Zemlyanskiy, Federico Lebron, and Sumit Sanghai.
\newblock {GQA}: Training generalized multi-query transformer models from multi-head checkpoints.
\newblock In \emph{Proceedings of the 2023 Conference on Empirical Methods in Natural Language Processing}, 2023.

\bibitem[Bai et~al.(2024)Bai, Lv, Zhang, Lyu, Tang, Huang, Du, Liu, Zeng, Hou, Dong, Tang, and Li]{bai2024longbench}
Yushi Bai, Xin Lv, Jiajie Zhang, Hongchang Lyu, Jiankai Tang, Zhidian Huang, Zhengxiao Du, Xiao Liu, Aohan Zeng, Lei Hou, Yuxiao Dong, Jie Tang, and Juanzi Li.
\newblock {LongBench}: A bilingual, multitask benchmark for long context understanding.
\newblock In \emph{Proceedings of the 62nd Annual Meeting of the Association for Computational Linguistics (Volume 1: Long Papers)}, 2024.

\bibitem[Bai et~al.(2026)Bai, Dong, Jiang, Lv, Du, Zeng, Tang, and Li]{bai2026indexcache}
Yushi Bai, Qian Dong, Ting Jiang, Xin Lv, Zhengxiao Du, Aohan Zeng, Jie Tang, and Juanzi Li.
\newblock Indexcache: Accelerating sparse attention via cross-layer index reuse.
\newblock In \emph{Third Conference on Language Modeling}, 2026.

\bibitem[Beltagy et~al.(2020)Beltagy, Peters, and Cohan]{beltagy2020longformer}
Iz~Beltagy, Matthew~E Peters, and Arman Cohan.
\newblock {Longformer}: The long-document transformer.
\newblock \emph{arXiv preprint arXiv:2004.05150}, 2020.

\bibitem[Burtenshaw et~al.(2026)Burtenshaw, {merve}, Cuenca, Gosthipaty, and Marafioti]{burtenshaw2026inkling}
Ben Burtenshaw, {merve}, Pedro Cuenca, Aritra~Roy Gosthipaty, and Andres Marafioti.
\newblock Welcome {Inkling} by {Thinking Machines}.
\newblock Hugging Face Blog, July 2026.
\newblock URL \url{https://huggingface.co/blog/thinkingmachines-inkling}.

\bibitem[Chen et~al.(2023)Chen, Wong, Chen, and Tian]{chen2023posinterpolation}
Shouyuan Chen, Sherman Wong, Liangjian Chen, and Yuandong Tian.
\newblock Extending context window of large language models via positional interpolation.
\newblock \emph{arXiv preprint arXiv:2306.15595}, 2023.

\bibitem[Chen et~al.(2025)Chen, Lv, Luan, Wang, and Liu]{chen2025hope}
Yuhan Chen, Ang Lv, Jian Luan, Bin Wang, and Wei Liu.
\newblock {HoPE}: A novel positional encoding without long-term decay for enhanced context awareness and extrapolation.
\newblock In \emph{Proceedings of the 63rd Annual Meeting of the Association for Computational Linguistics (Volume 1: Long Papers)}, 2025.

\bibitem[Child et~al.(2019)Child, Gray, Radford, and Sutskever]{child2019sparsetransformer}
Rewon Child, Scott Gray, Alec Radford, and Ilya Sutskever.
\newblock Generating long sequences with sparse transformers.
\newblock \emph{arXiv preprint arXiv:1904.10509}, 2019.

\bibitem[{DeepSeek-AI}(2024)]{deepseekai2024deepseekv2}
{DeepSeek-AI}.
\newblock {DeepSeek-V2}: A strong, economical, and efficient mixture-of-experts language model.
\newblock \emph{arXiv preprint arXiv:2405.04434}, 2024.

\bibitem[Ding et~al.(2024)Ding, Zhang, Zhang, Xu, Shang, Xu, Yang, and Yang]{ding2024longrope}
Yiran Ding, Li~Lyna Zhang, Chengruidong Zhang, Yuanyuan Xu, Ning Shang, Jiahang Xu, Fan Yang, and Mao Yang.
\newblock {L}ong{R}o{PE}: Extending {LLM} context window beyond 2 million tokens.
\newblock In \emph{Proceedings of the 41st International Conference on Machine Learning}, 2024.

\bibitem[Gelberg et~al.(2026)Gelberg, Eguchi, Akiba, and Cetin]{gelberg2026drope}
Yoav Gelberg, Koshi Eguchi, Takuya Akiba, and Edoardo Cetin.
\newblock Extending the context of pretrained llms by dropping their positional embedding.
\newblock In \emph{International Conference on Learning Representations}, 2026.

\bibitem[{GLM Team}(2026)]{glm53}
{GLM Team}.
\newblock {GLM-5.3}: Frontier coding with emergent cyber capabilities.
\newblock Z.ai Blog, August 2026.
\newblock URL \url{https://z.ai/blog/glm-5.3}.

\bibitem[Golovneva et~al.(2024)Golovneva, Wang, Weston, and Sukhbaatar]{golovneva2024cope}
Olga Golovneva, Tianlu Wang, Jason Weston, and Sainbayar Sukhbaatar.
\newblock Contextual position encoding: Learning to count what's important.
\newblock \emph{arXiv preprint arXiv:2405.18719}, 2024.

\bibitem[Golowich et~al.(2025)Golowich, Jelassi, Brandfonbrener, Kakade, and Malach]{golowich2025role}
Noah Golowich, Samy Jelassi, David Brandfonbrener, Sham~M. Kakade, and Eran Malach.
\newblock The role of sparsity for length generalization in {LLM}s.
\newblock In \emph{Proceedings of the 42nd International Conference on Machine Learning}, 2025.

\bibitem[Hsieh et~al.(2024)Hsieh, Sun, Kriman, Acharya, Rekesh, Jia, Zhang, and Ginsburg]{hsieh2024ruler}
Cheng-Ping Hsieh, Simeng Sun, Samuel Kriman, Shantanu Acharya, Dima Rekesh, Fei Jia, Yang Zhang, and Boris Ginsburg.
\newblock {RULER}: What's the real context size of your long-context language models?
\newblock In \emph{First Conference on Language Modeling}, 2024.

\bibitem[Hu et~al.(2026)Hu, Wei, Gu, Zhang, Liang, Li, Zhu, Wang, Han, Bai, et~al.]{hu2026hils}
Xiang Hu, Xinyu Wei, Hao Gu, Minshen Zhang, Tian Liang, Huayang Li, Lei Zhu, Yan Wang, Sirui Han, Yushi Bai, et~al.
\newblock Hierarchical sparse attention done right: Toward infinite context modeling.
\newblock \emph{arXiv preprint arXiv:2607.02980}, 2026.

\bibitem[Huang et~al.(2025)Huang, Li, He, Zhou, and Shechtman]{huang2025selfforcing}
Xun Huang, Zhengqi Li, Guande He, Mingyuan Zhou, and Eli Shechtman.
\newblock {Self Forcing}: Bridging the train-test gap in autoregressive video diffusion.
\newblock In \emph{Advances in Neural Information Processing Systems}, 2025.

\bibitem[Huang et~al.(2026)Huang, Gon{\c{c}}alves, Alvetreti, Li, Han, Ponti, Martins, and Treviso]{huang2026dashattention}
Yuxiang Huang, Nuno~MT Gon{\c{c}}alves, Federico Alvetreti, Lei Li, Xu~Han, Edoardo~M Ponti, Andr{\'e}~FT Martins, and Marcos~V Treviso.
\newblock Dashattention: Differentiable and adaptive sparse hierarchical attention.
\newblock \emph{arXiv preprint arXiv:2605.18753}, 2026.

\bibitem[Jiang et~al.(2024)Jiang, Li, Zhang, Wu, Luo, Ahn, Han, Abdi, Li, Lin, et~al.]{jiang2024minference}
Huiqiang Jiang, Yucheng Li, Chengruidong Zhang, Qianhui Wu, Xufang Luo, Surin Ahn, Zhenhua Han, Amir~H Abdi, Dongsheng Li, Chin-Yew Lin, et~al.
\newblock Minference 1.0: Accelerating pre-filling for long-context llms via dynamic sparse attention.
\newblock \emph{Advances in Neural Information Processing Systems}, 2024.

\bibitem[Kazemnejad et~al.(2023)Kazemnejad, Padhi, Natesan~Ramamurthy, Das, and Reddy]{kazemnejad2023impactnope}
Amirhossein Kazemnejad, Inkit Padhi, Karthikeyan Natesan~Ramamurthy, Payel Das, and Siva Reddy.
\newblock The impact of positional encoding on length generalization in transformers.
\newblock In \emph{Advances in Neural Information Processing Systems}, 2023.

\bibitem[{Kimi Team}(2025)]{kimiteam2025kimik2}
{Kimi Team}.
\newblock {Kimi K2}: Open agentic intelligence.
\newblock \emph{arXiv preprint arXiv:2507.20534}, 2025.

\bibitem[Lai et~al.(2026)Lai, Xu, Yang, Chen, Xu, Zeng, Li, Sun, Zhu, Zhang, et~al.]{lai2026msa}
Xunhao Lai, Weiqi Xu, Yufeng Yang, Qiaorui Chen, Yang Xu, Lunbin Zeng, Xiaolong Li, Haohai Sun, Haichao Zhu, Vito Zhang, et~al.
\newblock Minimax sparse attention.
\newblock \emph{arXiv preprint arXiv:2606.13392}, 2026.

\bibitem[Lee et~al.(2025)Lee, Park, Suh, and Hwang]{lee2025infinitehip}
Heejun Lee, Geon Park, Jaduk Suh, and Sung~Ju Hwang.
\newblock Infinitehip: Extending language model context up to 3 million tokens on a single gpu.
\newblock \emph{arXiv preprint arXiv:2502.08910}, 2025.

\bibitem[Li et~al.(2026{\natexlab{a}})Li, Zhao, Cai, and Sproat]{li2026repo}
Huayang Li, Tianyu Zhao, Deng Cai, and Richard Sproat.
\newblock {R}e{P}o: Language models with context re-positioning.
\newblock In \emph{Proceedings of the 43rd International Conference on Machine Learning}, 2026{\natexlab{a}}.

\bibitem[Li et~al.(2026{\natexlab{b}})Li, Zhu, Gu, Hu, Wang, Mi, Han, Liang, and Guo]{li2026sas}
Zhiwei Li, Lei Zhu, Hao Gu, Xiang Hu, Yan Wang, Haitao Mi, Sirui Han, Leo Liang, and Zhijiang Guo.
\newblock Sas: Simple attention sparsification via end-to-end optimization of context ranking.
\newblock \emph{arXiv preprint arXiv:2609.13141}, 2026{\natexlab{b}}.

\bibitem[Liu et~al.(2025)Liu, Mei, Lin, Xue, Wang, Xu, Wu, Zhang, Lin, Dong, et~al.]{liu2025deepseekv32-dsa}
Aixin Liu, Aoxue Mei, Bangcai Lin, Bing Xue, Bingxuan Wang, Bingzheng Xu, Bochao Wu, Bowei Zhang, Chaofan Lin, Chen Dong, et~al.
\newblock Deepseek-v3. 2: Pushing the frontier of open large language models.
\newblock \emph{arXiv preprint arXiv:2512.02556}, 2025.

\bibitem[Liu et~al.(2024{\natexlab{a}})Liu, Lin, Hewitt, Paranjape, Bevilacqua, Petroni, and Liang]{liu2024lostinthemiddle}
Nelson~F Liu, Kevin Lin, John Hewitt, Ashwin Paranjape, Michele Bevilacqua, Fabio Petroni, and Percy Liang.
\newblock Lost in the middle: How language models use long contexts.
\newblock \emph{Transactions of the Association for Computational Linguistics}, 12:\penalty0 157--173, 2024{\natexlab{a}}.

\bibitem[Liu et~al.(2024{\natexlab{b}})Liu, Yan, An, Qiu, and Lin]{liu2024scalinglawsrope}
Xiaoran Liu, Hang Yan, Chenxin An, Xipeng Qiu, and Dahua Lin.
\newblock Scaling laws of {RoPE}-based extrapolation.
\newblock In \emph{International Conference on Learning Representations}, 2024{\natexlab{b}}.

\bibitem[Liu et~al.(2021)Liu, Lin, Cao, Hu, Wei, Zhang, Lin, and Guo]{liu2021swin}
Ze~Liu, Yutong Lin, Yue Cao, Han Hu, Yixuan Wei, Zheng Zhang, Stephen Lin, and Baining Guo.
\newblock Swin transformer: Hierarchical vision transformer using shifted windows.
\newblock In \emph{2021 IEEE/CVF international conference on computer vision (ICCV)}, 2021.

\bibitem[Lu et~al.(2026)Lu, Jiang, Liu, Du, Jiang, Hong, Liu, He, Yuan, Wang, et~al.]{lu2026moba}
Enzhe Lu, Zhejun Jiang, Jingyuan Liu, Yulun Du, Tao Jiang, Chao Hong, Shaowei Liu, Weiran He, Enming Yuan, Yuzhi Wang, et~al.
\newblock Moba: Mixture of block attention for long-context llms.
\newblock \emph{Advances in Neural Information Processing Systems}, 2026.

\bibitem[Ma et~al.(2025)Ma, Zheng, Shi, Zhao, Jia, Huang, Lin, Li, Yang, Peng, Zhang, and Liu]{ma2025veomni}
Qianli Ma, Yaowei Zheng, Zhelun Shi, Zhongkai Zhao, Bin Jia, Ziyue Huang, Zhiqi Lin, Youjie Li, Jiacheng Yang, Yanghua Peng, Zhi Zhang, and Xin Liu.
\newblock {VeOmni}: Scaling any modality model training with model-centric distributed recipe zoo.
\newblock \emph{arXiv preprint arXiv:2508.02317}, 2025.

\bibitem[{Mistral AI}(2026)]{mistralai2026mistrallarge4}
{Mistral AI}.
\newblock {Mistral Large 4}.
\newblock Technical report, Mistral AI, 2026.
\newblock URL \url{https://huggingface.co/mistralai/Mistral-Large-4-1T-A52B/blob/main/assets/technical_report.pdf}.

\bibitem[Nakanishi(2025)]{nakanishi2025ssmax}
Ken~M Nakanishi.
\newblock Scalable-softmax is superior for attention.
\newblock \emph{arXiv preprint arXiv:2501.19399}, 2025.

\bibitem[Peng et~al.(2024)Peng, Quesnelle, Fan, and Shippole]{peng2024yarn}
Bowen Peng, Jeffrey Quesnelle, Honglu Fan, and Enrico Shippole.
\newblock Yarn: Efficient context window extension of large language models.
\newblock In \emph{International Conference on Learning Representations}, 2024.

\bibitem[Press et~al.(2022)Press, Smith, and Lewis]{press2021alibi}
Ofir Press, Noah Smith, and Mike Lewis.
\newblock Train short, test long: Attention with linear biases enables input length extrapolation.
\newblock In \emph{International Conference on Learning Representations}, 2022.

\bibitem[Qiu et~al.(2026)Qiu, Wang, Li, Li, Xu, Wang, Zhang, Men, Mao, Zhang, et~al.]{qiu2026qwen38next-qsa}
Zihan Qiu, Zekun Wang, Xiao Li, Yanpeng Li, Yang Xu, Yixuan Wang, Huaqing Zhang, Rui Men, Bochao Mao, Chengruidong Zhang, et~al.
\newblock On the design of qwen3. 8-next architecture: Evaluation, efficiency, and training stability.
\newblock \emph{arXiv preprint arXiv:2608.30320}, 2026.

\bibitem[Ribar et~al.(2024)Ribar, Chelombiev, Hudlass-Galley, Blake, Luschi, and Orr]{ribar2023sparq}
Luka Ribar, Ivan Chelombiev, Luke Hudlass-Galley, Charlie Blake, Carlo Luschi, and Douglas Orr.
\newblock {SparQ} attention: Bandwidth-efficient {LLM} inference.
\newblock In \emph{Proceedings of the 41st International Conference on Machine Learning}, 2024.

\bibitem[Ruoss et~al.(2023)Ruoss, Del{\'e}tang, Genewein, Grau-Moya, Csord{\'a}s, Bennani, Legg, and Veness]{ruoss2023randomized}
Anian Ruoss, Gr{\'e}goire Del{\'e}tang, Tim Genewein, Jordi Grau-Moya, R{\'o}bert Csord{\'a}s, Mehdi Bennani, Shane Legg, and Joel Veness.
\newblock Randomized positional encodings boost length generalization of transformers.
\newblock In \emph{Proceedings of the 61st Annual Meeting of the Association for Computational Linguistics (Volume 2: Short Papers)}, 2023.

\bibitem[Shang et~al.(2025)Shang, Zhang, Wang, Zhang, Lopez, Yang, Chen, and Yang]{shang2025longrope2}
Ning Shang, Li~Lyna Zhang, Siyuan Wang, Gaokai Zhang, Gilsinia Lopez, Fan Yang, Weizhu Chen, and Mao Yang.
\newblock {L}ong{R}o{PE}2: Near-lossless {LLM} context window scaling.
\newblock In \emph{Proceedings of the 42nd International Conference on Machine Learning}, 2025.

\bibitem[Shazeer(2019)]{shazeer2019multiquery}
Noam Shazeer.
\newblock Fast transformer decoding: One write-head is all you need.
\newblock \emph{arXiv preprint arXiv:1911.02150}, 2019.

\bibitem[Su et~al.(2024)Su, Ahmed, Lu, Pan, Wen, and Liu]{su2024roformer}
Jianlin Su, Murtadha Ahmed, Yu~Lu, Shengfeng Pan, Bo~Wen, and Yunfeng Liu.
\newblock {RoFormer}: Enhanced transformer with rotary position embedding.
\newblock \emph{Neurocomputing}, 568:\penalty0 127063, 2024.

\bibitem[Tan et~al.(2025)Tan, Yang, Courville, Panda, and Shen]{tan2025stickbreakingattn}
Shawn Tan, Songlin Yang, Aaron Courville, Rameswar Panda, and Yikang Shen.
\newblock Scaling stick-breaking attention: An efficient implementation and in-depth study.
\newblock In \emph{International Conference on Learning Representations}, 2025.

\bibitem[Tang et~al.(2024)Tang, Zhao, Zhu, Xiao, Kasikci, and Han]{xiao2024quest}
Jiaming Tang, Yilong Zhao, Kan Zhu, Guangxuan Xiao, Baris Kasikci, and Song Han.
\newblock {QUEST}: Query-aware sparsity for efficient long-context {LLM} inference.
\newblock In \emph{Proceedings of the 41st International Conference on Machine Learning}, 2024.

\bibitem[Team et~al.(2026)Team, Abd, Aggarwal, Algayres, Andreev, Bachem, Ballantyne, Brick, C{\u{a}}rbune, Casbon, et~al.]{team2026gemma4}
Gemma Team, Sherif~El Abd, Vaibhav Aggarwal, Robin Algayres, Alek Andreev, Olivier Bachem, Ian Ballantyne, Cormac Brick, Victor C{\u{a}}rbune, Michelle Casbon, et~al.
\newblock Gemma 4 technical report.
\newblock \emph{arXiv preprint arXiv:2607.02770}, 2026.

\bibitem[{Team Olmo} et~al.(2025){Team Olmo}, Ettinger, Bertsch, Kuehl, Graham, Heineman, Groeneveld, Brahman, Timbers, Ivison, Morrison, Poznanski, Lo, Soldaini, Jordan, Chen, Noukhovitch, Lambert, Walsh, Dasigi, Berry, Malik, Shah, Geng, Arora, Gupta, Anderson, Xiao, Murray, Romero, Graf, Asai, Bhagia, Wettig, Liu, Rangapur, Anastasiades, Huang, Schwenk, Trivedi, Magnusson, Lochner, Liu, Miranda, Sap, Morgan, Schmitz, Guerquin, Wilson, Huff, Bras, Xin, Shao, Skjonsberg, Shen, Li, Wilde, Pyatkin, Merrill, Chang, Gu, Zeng, Sabharwal, Zettlemoyer, Koh, Farhadi, Smith, and Hajishirzi]{olmo3}
{Team Olmo}, Allyson Ettinger, Amanda Bertsch, Bailey Kuehl, David Graham, David Heineman, Dirk Groeneveld, Faeze Brahman, Finbarr Timbers, Hamish Ivison, Jacob Morrison, Jake Poznanski, Kyle Lo, Luca Soldaini, Matt Jordan, Mayee Chen, Michael Noukhovitch, Nathan Lambert, Pete Walsh, Pradeep Dasigi, Robert Berry, Saumya Malik, Saurabh Shah, Scott Geng, Shane Arora, Shashank Gupta, Taira Anderson, Teng Xiao, Tyler Murray, Tyler Romero, Victoria Graf, Akari Asai, Akshita Bhagia, Alexander Wettig, Alisa Liu, Aman Rangapur, Chloe Anastasiades, Costa Huang, Dustin Schwenk, Harsh Trivedi, Ian Magnusson, Jaron Lochner, Jiacheng Liu, Lester James~V. Miranda, Maarten Sap, Malia Morgan, Michael Schmitz, Michal Guerquin, Michael Wilson, Regan Huff, Ronan~Le Bras, Rui Xin, Rulin Shao, Sam Skjonsberg, Shannon~Zejiang Shen, Shuyue~Stella Li, Tucker Wilde, Valentina Pyatkin, Will Merrill, Yapei Chang, Yuling Gu, Zhiyuan Zeng, Ashish Sabharwal, Luke Zettlemoyer, Pang~Wei Koh, Ali Farhadi, Noah~A. Smith, and Hannaneh Hajishirzi.
\newblock {OLMo 3}.
\newblock \emph{arXiv preprint arXiv:2512.13961}, 2025.

\bibitem[Vaswani et~al.(2017)Vaswani, Shazeer, Parmar, Uszkoreit, Jones, Gomez, Kaiser, and Polosukhin]{vaswani2017attention}
Ashish Vaswani, Noam Shazeer, Niki Parmar, Jakob Uszkoreit, Llion Jones, Aidan~N Gomez, {\L}ukasz Kaiser, and Illia Polosukhin.
\newblock Attention is all you need.
\newblock In \emph{Advances in Neural Information Processing Systems}, 2017.

\bibitem[Veisi et~al.(2025)Veisi, Fartoot, and Amirzadeh]{veisi2025carope}
Ali Veisi, Delaram Fartoot, and Hamidreza Amirzadeh.
\newblock Context-aware rotary position embedding.
\newblock \emph{arXiv preprint arXiv:2507.23083}, 2025.

\bibitem[Xiao et~al.(2024{\natexlab{a}})Xiao, Zhang, Han, Xiao, Lin, Zhang, Liu, and Sun]{xiao2024infllm}
Chaojun Xiao, Pengle Zhang, Xu~Han, Guangxuan Xiao, Yankai Lin, Zhengyan Zhang, Zhiyuan Liu, and Maosong Sun.
\newblock Infllm: Training-free long-context extrapolation for llms with an efficient context memory.
\newblock \emph{Advances in neural information processing systems}, 2024{\natexlab{a}}.

\bibitem[Xiao et~al.(2024{\natexlab{b}})Xiao, Tian, Chen, Han, and Lewis]{xiao2024streamingllm}
Guangxuan Xiao, Yuandong Tian, Beidi Chen, Song Han, and Mike Lewis.
\newblock Efficient streaming language models with attention sinks.
\newblock In \emph{International Conference on Learning Representations}, 2024{\natexlab{b}}.

\bibitem[Xiao et~al.(2025)Xiao, Tang, Zuo, Guo, Yang, Tang, Fu, and Han]{xiao2024duoattention}
Guangxuan Xiao, Jiaming Tang, Jingwei Zuo, Junxian Guo, Shang Yang, Haotian Tang, Yao Fu, and Song Han.
\newblock {DuoAttention}: Efficient long-context {LLM} inference with retrieval and streaming heads.
\newblock In \emph{International Conference on Learning Representations}, 2025.

\bibitem[Xiong et~al.(2024)Xiong, Liu, Molybog, Zhang, Bhargava, Hou, Martin, Rungta, Sankararaman, Oguz, et~al.]{xiong2023effectivelongcontext}
Wenhan Xiong, Jingyu Liu, Igor Molybog, Hejia Zhang, Prajjwal Bhargava, Rui Hou, Louis Martin, Rashi Rungta, Karthik~Abinav Sankararaman, Barlas Oguz, et~al.
\newblock Effective long-context scaling of foundation models.
\newblock In \emph{Proceedings of the 2024 Conference of the North American Chapter of the Association for Computational Linguistics: Human Language Technologies (Volume 1: Long Papers)}, 2024.

\bibitem[Xu et~al.(2026)Xu, Lin, Xue, Wang, Xu, Wu, Zhang, Lin, Dong, Ling, et~al.]{xu2026deepseekv4-dsa}
Anyi Xu, Bangcai Lin, Bing Xue, Bingxuan Wang, Bingzheng Xu, Bochao Wu, Bowei Zhang, Chaofan Lin, Chen Dong, Chenchen Ling, et~al.
\newblock Deepseek-v4: Towards highly efficient million-token context intelligence.
\newblock \emph{arXiv preprint arXiv:2606.19348}, 2026.

\bibitem[Yen et~al.(2025)Yen, Gao, Hou, Ding, Fleischer, Izsak, Wasserblat, and Chen]{yen2024helmet}
Howard Yen, Tianyu Gao, Minmin Hou, Ke~Ding, Daniel Fleischer, Peter Izsak, Moshe Wasserblat, and Danqi Chen.
\newblock {HELMET}: How to evaluate long-context language models effectively and thoroughly.
\newblock In \emph{International Conference on Learning Representations}, 2025.

\bibitem[Yin et~al.(2025)Yin, Zhang, Zhang, Freeman, Durand, Shechtman, and Huang]{yin2024causvid}
Tianwei Yin, Qiang Zhang, Richard Zhang, William~T. Freeman, Fr{\'e}do Durand, Eli Shechtman, and Xun Huang.
\newblock From slow bidirectional to fast autoregressive video diffusion models.
\newblock In \emph{Proceedings of the IEEE/CVF Conference on Computer Vision and Pattern Recognition}, 2025.

\bibitem[Yuan et~al.(2025)Yuan, Gao, Dai, Luo, Zhao, Zhang, Xie, Wei, Wang, Xiao, et~al.]{yuan2025nsa}
Jingyang Yuan, Huazuo Gao, Damai Dai, Junyu Luo, Liang Zhao, Zhengyan Zhang, Zhenda Xie, Yuxing Wei, Lean Wang, Zhiping Xiao, et~al.
\newblock Native sparse attention: Hardware-aligned and natively trainable sparse attention.
\newblock In \emph{Proceedings of the 63rd Annual Meeting of the Association for Computational Linguistics (Volume 1: Long Papers)}, 2025.

\bibitem[Zaheer et~al.(2020)Zaheer, Guruganesh, Dubey, Ainslie, Alberti, Ontanon, Pham, Ravula, Wang, Yang, et~al.]{zaheer2020bigbird}
Manzil Zaheer, Guru Guruganesh, Kumar~Avinava Dubey, Joshua Ainslie, Chris Alberti, Santiago Ontanon, Philip Pham, Anirudh Ravula, Qifan Wang, Li~Yang, et~al.
\newblock {Big Bird}: Transformers for longer sequences.
\newblock In \emph{Advances in Neural Information Processing Systems}, 2020.

\bibitem[Zhang et~al.(2023)Zhang, Sheng, Zhou, Chen, Zheng, Cai, Song, Tian, R{\'e}, Barrett, et~al.]{zhang2023h2o}
Zhenyu Zhang, Ying Sheng, Tianyi Zhou, Tianlong Chen, Lianmin Zheng, Ruisi Cai, Zhao Song, Yuandong Tian, Christopher R{\'e}, Clark Barrett, et~al.
\newblock {H2O}: Heavy-hitter oracle for efficient generative inference of large language models.
\newblock In \emph{Advances in Neural Information Processing Systems}, 2023.

\bibitem[Zhu et~al.(2023)Zhu, Wang, Ke, Zhang, and Lau]{zhu2023biformer}
Lei Zhu, Xinjiang Wang, Zhanghan Ke, Wayne Zhang, and Rynson~WH Lau.
\newblock Biformer: Vision transformer with bi-level routing attention.
\newblock In \emph{Proceedings of the IEEE/CVF conference on computer vision and pattern recognition}, 2023.

\end{thebibliography}
